\documentclass{article}

\usepackage[nonatbib,final]{neurips_2020}

\usepackage[utf8]{inputenc} 
\usepackage[T1]{fontenc}    
\usepackage{hyperref}       
\usepackage{url}            
\usepackage{booktabs}       
\usepackage{amsfonts}       
\usepackage{nicefrac}       
\usepackage{microtype}      

\usepackage{amsmath,amssymb}       
\usepackage{bm}      
\usepackage[pdftex]{graphicx}
\usepackage{algorithm}
\usepackage[noend]{algpseudocode}
\usepackage{xspace}
\usepackage{etoc}
\usepackage{cite}

\makeatletter
\def\algbackskip{\hskip-\ALG@thistlm}
\makeatother

\usepackage{color}

\newcommand{\ignore}[1]{}

\newcommand{\x}{\bm{\theta}}

\newcommand{\xx}{\bm{\Theta}}
\newcommand{\y}{\bm{y}}
\newcommand{\z}{\bm{z}}
\newcommand{\mmu}{\bm{\mu}}
\newcommand{\vmu}{\bm{\mu}}

\newcommand{\vlambda}{\bm{\lambda}}

\newcommand{\X}{\mathcal{X}}

\newcommand{\nparams}{D}

\newcommand{\supplement}{Supplement\xspace}
\newcommand{\LL}{log-likelihood\xspace}

\newcommand{\normpdf}[3]{\mathcal{N}\left({#1}; {#2}, {#3 } \right)}

\newcommand{\data}{\mathcal{D}}
\newcommand{\vtheta}{\bm{\theta}}

\newcommand{\like}{p(\data|\x)}
\newcommand{\post}{p(\x|\data)}
\newcommand{\ev}{p(\data)}

\newcommand{\qparams}{\bm{\phi}}

\newcommand{\qp}{q_{\qparams}}
\newcommand{\qx}{q(\x)}
\newcommand{\qpx}{q_{\qparams}(\x)}

\newcommand{\vgp}{\bm{\psi}}

\newcommand{\gpdata}{{\bm{\Xi}}}

\newcommand{\covar}{C}
\newcommand{\mat}[1]{\text{\bf{#1}}}
\newcommand{\logjoint}{\mathcal{G}}

\newcommand{\sigman}{\sigma_\text{obs}}
\newcommand{\ddx}{{z}}
\newcommand{\att}{\beta}

\newcommand{\VBMC}{\textsc{vbmc}\xspace}
\newcommand{\vbmc}{\textsc{vbmc}\xspace}
\newcommand{\wsabi}{\textsc{wsabi}\xspace}
\newcommand{\wsabil}{\textsc{wsabi-l}\xspace}
\newcommand{\gpimiqr}{\textsc{gp-imiqr}\xspace}
\newcommand{\MCMC}{\textsc{mcmc}\xspace}
\newcommand{\MAP}{MAP\xspace}

\newcommand{\plb}{\texttt{PLB}\xspace}
\newcommand{\pub}{\texttt{PUB}\xspace}
\newcommand{\lb}{\texttt{LB}\xspace}
\newcommand{\ub}{\texttt{UB}\xspace}

\newcommand{\npro}{\textsc{npro}\xspace}
\newcommand{\VIQR}{\textsc{viqr}\xspace}
\newcommand{\viqr}{\textsc{viqr}\xspace}
\newcommand{\IMIQR}{\textsc{imiqr}\xspace}
\newcommand{\imiqr}{\textsc{imiqr}\xspace}
\newcommand{\EIG}{\textsc{eig}\xspace}
\newcommand{\eig}{\textsc{eig}\xspace}
\newcommand{\iqr}{\textsc{iqr}\xspace}

\algnewcommand\algorithmicinput{\textbf{Input:}}
\algnewcommand\INPUT{\item[\algorithmicinput]}

\algloop{Do}

\title{Variational Bayesian Monte Carlo \\ with Noisy Likelihoods}

\author{
Luigi Acerbi\thanks{Previous affiliation: Department of Basic Neuroscience, University of Geneva.}
\\
Department of Computer Science \\
University of Helsinki \\
\texttt{luigi.acerbi@helsinki.fi}
}

\begin{document}

\etocdepthtag.toc{mtchapter}
\etocsettagdepth{mtchapter}{subsubsection}
\etocsettagdepth{mtappendix}{none}

\maketitle
\setcounter{footnote}{0}

\vspace{-1em}
\begin{abstract}
\vspace{-0.25em}

Variational Bayesian Monte Carlo (\VBMC) is a recently introduced framework that uses Gaussian process surrogates to perform approximate Bayesian inference in models with \emph{black-box}, non-cheap likelihoods. In this work, we extend \VBMC to deal with noisy log-likelihood evaluations, such as those arising from simulation-based models. We introduce new `global' acquisition functions, such as \emph{expected information gain} (\EIG) and \emph{variational interquantile range} (\VIQR), which are robust to noise and can be efficiently evaluated within the \VBMC setting. In a novel, challenging, noisy-inference benchmark comprising of a variety of models with real datasets from computational and cognitive neuroscience, \VBMC+\VIQR achieves state-of-the-art performance in recovering the ground-truth posteriors and model evidence.
In particular, our method vastly outperforms `local' acquisition functions and other surrogate-based inference methods while keeping a small algorithmic cost. Our benchmark corroborates \VBMC as a general-purpose technique for sample-efficient black-box Bayesian inference also with noisy models.
\end{abstract}

\vspace{-0.5em}
\section{Introduction}

\vspace{-0.25em}






Bayesian inference provides a principled framework for uncertainty quantification and model selection via computation of the posterior distribution over model parameters and of the model evidence \cite{mackay2003information,gelman2013bayesian}. 
However, for many \emph{black-box} models of interest in fields such as computational biology and neuroscience, (log-)likelihood evaluations are computationally expensive (thus limited in number) 
and noisy due to, e.g., simulation-based approximations \cite{wood2010statistical,price2018bayesian}.
These features make standard techniques for approximate Bayesian inference such as Markov Chain Monte Carlo (\MCMC) ineffective.

Variational Bayesian Monte Carlo (\vbmc) is a recently proposed framework for Bayesian inference with non-cheap models \cite{acerbi2018variational,acerbi2019exploration}. 
\VBMC performs variational inference using a Gaussian process (GP \cite{rasmussen2006gaussian}) as a statistical surrogate model for the expensive log posterior distribution. The GP model is refined via active sampling, guided by a `smart' \emph{acquisition function} that exploits uncertainty and other features of the surrogate. \vbmc is particularly efficient thanks to a representation that affords fast integration via Bayesian quadrature \cite{ohagan1991bayes,ghahramani2003bayesian}, and unlike other surrogate-based techniques it performs both posterior and model inference \cite{acerbi2018variational}.
However, the original formulation of \vbmc does not support noisy model evaluations, and recent work has shown that surrogate-based approaches that work well in the noiseless case may fail in the presence of even small amounts of noise \cite{jarvenpaa2020parallel}.

In this work, we extend \vbmc to deal robustly and effectively with noisy log-likelihood evaluations, broadening the class of models that can be estimated via the method. 
With our novel contributions, \VBMC outperforms other state-of-the-art surrogate-based techniques for black-box Bayesian inference in the presence of noisy evaluations -- in terms of  speed, robustness and quality of solutions.

\vspace{-0.5em}
\paragraph*{Contributions}

We make the following contributions: (1) we introduce several new acquisition functions for \VBMC that explicitly account for noisy log-likelihood evaluations, and leverage the variational representation to achieve much faster evaluation than competing methods; (2) we introduce \emph{variational whitening}, a technique to deal with non-axis aligned posteriors, which are otherwise potentially problematic for \VBMC (and GP surrogates more in general) in the presence of noise; (3) we build a novel and challenging noisy-inference benchmark that includes five different models from computational and cognitive neuroscience, ranging from $3$ to $9$ parameters, and applied to real datasets, in which we test \VBMC and other state-of-the-art surrogate-based inference techniques.
The new features have been implemented in \VBMC: \url{https://github.com/lacerbi/vbmc}.


\vspace{-0.5em}
\paragraph*{Related work}

Our paper extends the \vbmc framework \cite{acerbi2018variational,acerbi2019exploration} by building on recent information-theoretical approaches to adaptive Bayesian quadrature \cite{gessner2019active}, and on recent theoretical and empirical results for GP-surrogate Bayesian inference for simulation-based models \cite{jarvenpaa2018gaussian,jarvenpaa2019efficient,jarvenpaa2020parallel}. For noiseless evaluations, previous work has used GP surrogates for estimation of posterior distributions \cite{rasmussen2003gaussian,kandasamy2015bayesian,wang2017adaptive} and Bayesian quadrature for calculation of the model evidence \cite{ghahramani2003bayesian,osborne2012active,gunter2014sampling, briol2015frank,chai2019automated}.
Our method is also closely related to (noisy) Bayesian optimization \cite{jones1998efficient,brochu2010tutorial,snoek2012practical,picheny2013quantile,gutmann2016bayesian,acerbi2017practical,letham2019constrained}. A completely different approach, but worth mentioning for the similar goal, trains deep networks on simulated data to reconstruct approximate Bayesian posteriors from data or summary statistics thereof \cite{papamakarios2016fast,lueckmann2017flexible,greenberg2019automatic,gonccalves2019training}.

\vspace{-0.35em}
\section{Variational Bayesian Monte Carlo (\VBMC)}
\label{sec:vbmc}

We summarize here the Variational Bayesian Monte Carlo (\VBMC) framework \cite{acerbi2018variational}. 
If needed, we refer the reader to the \supplement for a recap of key concepts in variational inference, GPs and Bayesian quadrature.
Let $f = \log \like p(\x)$ be the \emph{target} log joint probability (unnormalized posterior), where $\like$ is the model likelihood for dataset $\data$ and parameter vector $\x \in \X \subseteq \mathbb{R}^\nparams$, and $p(\x)$ the prior. We assume that only a limited number of log-likelihood evaluations are available, up to several hundreds. 
\VBMC works by iteratively improving a variational approximation $q_{\qparams}(\x)$, indexed by $\qparams$, of the true posterior density. In each iteration $t$, the algorithm: \vspace{-0.15em}
\begin{enumerate}
\item Actively samples sequentially $n_\text{active}$ `promising' new points, by iteratively maximizing a given acquisition function $a(\x): \X \rightarrow \mathbb{R}$;
for each selected point $\x_\star$ evaluates the target $\y_\star \equiv f(\x_\star)$ ($n_\text{active} = 5$ by default). \vspace{-0.15em}
\item Trains a GP surrogate model of the log joint $f$, given the training set $\gpdata_t = \left\{\xx_t, \y_t \right\}$ of input points and their associated observed values so far. \vspace{-0.15em}
\item Updates the variational posterior parameters $\qparams_t$ by optimizing the surrogate ELBO (variational lower bound on the model evidence) calculated via Bayesian quadrature. \vspace{-0.15em}
\end{enumerate}
This loop repeats until reaching a termination criterion (e.g., budget of function evaluations or lack of improvement over several iterations), and the algorithm returns both the variational posterior and posterior mean and variance of the ELBO. 
\VBMC includes an initial \emph{warm-up} stage to converge faster to regions of high posterior probability, before starting to refine the variational solution (see \cite{acerbi2018variational}).

\vspace{-0.25em}
\subsection{Basic features}

We briefly describe here basic features of the original \VBMC framework \cite{acerbi2018variational} (see also \supplement).

\vspace{-0.5em}
\paragraph{Variational posterior}
The variational posterior is a flexible mixture of $K$ multivariate Gaussians, 
$q(\x) \equiv q_{\qparams}(\x) = \sum_{k = 1}^K w_k \normpdf{\x}{\mmu_k}{\sigma_k^2 \mathbf{\Sigma}}$,
where $w_k$, $\mmu_k$, and $\sigma_k$ are, respectively, the mixture weight, mean, and scale of the $k$-th component; and $\mathbf{\Sigma}$  is a common diagonal covariance matrix $\mathbf{\Sigma} \equiv \text{diag}[{{\lambda}^{(1)}}^2,\ldots,{\lambda^{(\nparams)}}^2]$.
For a given $K$, the variational parameter vector is $\qparams \equiv (w_1,\ldots,w_K,$ $\vmu_1, \ldots, \vmu_K,$ $\sigma_1, \ldots, \sigma_K, \vlambda)$. $K$ is set adaptively; fixed to $K = 2$ during warm-up, and then increasing each iteration if it leads to an improvement of the ELBO.

\vspace{-0.5em}
\paragraph{Gaussian process model}
In \VBMC, the log joint $f$ is approximated by a GP surrogate model with a squared exponential (rescaled Gaussian) kernel, a Gaussian likelihood, 
and a \emph{negative quadratic} mean function which ensures finiteness of the variational objective \cite{acerbi2018variational,acerbi2019exploration}. In the original formulation, observations are assumed to be exact (non-noisy), so the GP likelihood only included a small observation noise $\sigma^2_\text{obs}$ for numerical stability \cite{gramacy2012cases}.
GP hyperparameters are estimated initially via \MCMC sampling \cite{neal2003slice}, when there is larger uncertainty about the GP model, and later via a maximum-a-posteriori (\MAP) estimate using gradient-based optimization (see \cite{acerbi2018variational} for details).

\vspace{-0.5em}
\paragraph{The Evidence Lower Bound (ELBO)}
Using the GP surrogate model $f$, and for a given variational posterior $q_{\qparams}$, the posterior mean of the surrogate ELBO can be estimated as
\begin{equation} \label{eq:elbo}
\mathbb{E}_{f | \gpdata}\left[\text{ELBO}(\qparams)\right] =  \mathbb{E}_{f | \gpdata}\left[\mathbb{E}_{\qparams} \left[f \right]\right] + \mathcal{H}[q_{\qparams}],
\end{equation}
where $\mathbb{E}_{f | \gpdata}\left[\mathbb{E}_{\qparams} \left[f \right]\right]$ is the posterior mean of the expected log joint under the GP model, and $\mathcal{H}[q_{\qparams}]$ is the entropy of the variational posterior. In particular, the expected log joint $\logjoint$ takes the form
\begin{equation} \label{eq:logjoint}
\logjoint\left[q_{\qparams}| f\right] \equiv \mathbb{E}_{\qparams}\left[ f \right] = \int q_{\qparams}(\x) f(\x) d \x.
\end{equation} 
Crucially, the choice of variational family and GP representation affords closed-form solutions for the posterior mean and variance of Eq. \ref{eq:logjoint} (and of their gradients) by means of Bayesian quadrature \cite{ohagan1991bayes,ghahramani2003bayesian}. The entropy of $q_{\qparams}$ and its gradient are estimated via simple Monte Carlo and the reparameterization trick \cite{kingma2013auto,miller2016variational}, such that  Eq. \ref{eq:elbo} can be optimized via stochastic gradient ascent \cite{kingma2014adam}.

\vspace{-0.5em}
\paragraph{Acquisition function}
During the active sampling stage, new points to evaluate are chosen sequentially by maximizing a given \emph{acquisition function} $a(\x): \X \rightarrow \mathbb{R}$ constructed to represent useful search heuristics \cite{kanagawa2019convergence}. 
The \VBMC paper introduced \emph{prospective uncertainty sampling} \cite{acerbi2018variational}, 
\begin{equation} \label{eq:acqpro}
a_\text{pro}(\x) = s^2_{\gpdata}(\x) q_{\qparams}(\x) \exp\left( \overline{f}_{\gpdata}(\x)\right),
\end{equation}
where $\overline{f}_{\gpdata}(\x)$ and $s^2_{\gpdata}(\x)$ are, respectively, the GP posterior latent mean and variance at $\x$ given the current training set $\gpdata$. 
Effectively, $a_\text{pro}$ promotes selection of new points from regions of high probability density, as represented by the variational posterior and (exponentiated) posterior mean of the surrogate log-joint, for which we are also highly uncertain (high variance of the GP surrogate).


\vspace{-0.5em}
\paragraph{Inference space} The variational posterior and GP surrogate in \VBMC are defined in an unbounded \emph{inference space} equal to $\mathbb{R}^\nparams$. Parameters that are subject to bound constraints are mapped to the inference space via a shifted and rescaled logit transform, with an appropriate Jacobian correction to the log-joint. Solutions are transformed back to the original space via a matched inverse transform, e.g., a shifted and rescaled logistic function for bound parameters (see \cite{carpenter2016stan,acerbi2018variational}).

\vspace{-0.25em}
\subsection{Variational whitening} 

One issue of the standard \VBMC representation of both the variational posterior and GP surrogate is that it is axis-aligned, which makes it ill-suited to deal with highly correlated posteriors. As a simple and inexpensive solution, we introduce here \emph{variational whitening}, which consists of a linear transformation $\mat{W}$ of the inference space (a rotation and rescaling) such that the variational posterior $q_{\qparams}$ obtains unit diagonal covariance matrix. Since $q_{\qparams}$ is a mixture of Gaussians in inference space, its covariance matrix $\mat{C}_{\qparams}$ is available in closed form and we can calculate the whitening transform $\mat{W}$ by performing a singular value decomposition (SVD) of $\mat{C}_{\qparams}$.
We start performing variational whitening a few iterations after the end of warm-up, and then at increasingly more distant intervals. By default we use variational whitening with all variants of \VBMC tested in this paper; see the \supplement for an ablation study demonstrating its usefulness and for further implementation details.

\vspace{-0.25em}
\section{\VBMC with noisy likelihood evaluations}
\label{sec:noisy}

Extending the framework described in Section \ref{sec:vbmc}, we now assume that evaluations of the \LL $y_n$ can be noisy, that is
\begin{equation} \label{eq:obsmodel}
y_n = f(\x_n) + \sigman(\x_n) \varepsilon_n, \qquad \varepsilon_n \stackrel{\text{i.i.d.}}{\sim} \mathcal{N}\left(0,1\right),
\end{equation}
where $\sigman : \X \rightarrow [\sigma_\text{min},\infty)$ is a function of the input space that determines the standard deviation (SD) of the observation noise. For this work, we use $\sigma_\text{min}^2 = 10^{-5}$ and we assume that the evaluation of the log-likelihood at $\x_n$ returns both $y_n$ and a reasonable estimate $(\widehat{\sigma}_\text{obs})_n$ of $\sigman(\x_n)$. Here we estimate $\sigman(\x)$ outside the training set via a nearest-neighbor approximation (see \supplement), but more sophisticated methods could be used (e.g., by training a GP model on $\sigman(\x_n)$ \cite{ankenman2010stochastic}).

The \emph{synthetic likelihood} (SL) technique \cite{wood2010statistical,price2018bayesian} and \emph{inverse binomial sampling} (IBS) \cite{haldane1945method,van2020unbiased} are examples of log-likelihood estimation methods for simulation-based models that satisfy the assumptions of our observation model (Eq. \ref{eq:obsmodel}). Recent work demonstrated empirically that log-SL estimates are approximately normally distributed, and their SD can be estimated accurately via bootstrap \cite{jarvenpaa2020parallel}. IBS is a recently reintroduced statistical technique that produces both normally-distributed, unbiased estimates of the log-likelihood and calibrated estimates of their variance \cite{van2020unbiased}.

\begin{figure}[htb]
  \includegraphics[width=\linewidth]{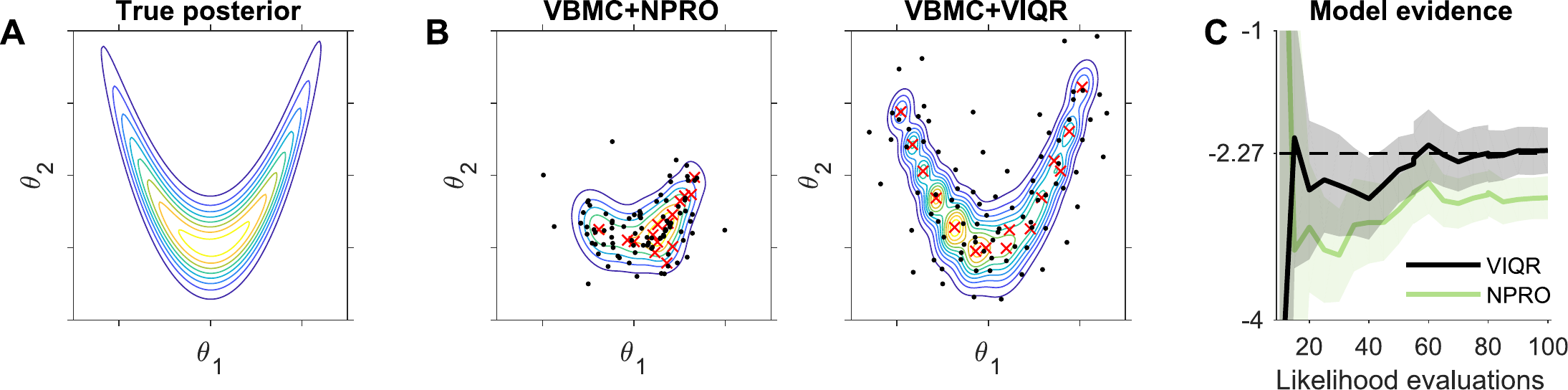}  
\vspace{-1.75em}
  \caption{{\bf \VBMC with noisy likelihoods.} \textbf{A.} True target pdf ($\nparams = 2$). We assume noisy log-likelihood evaluations with $\sigman = 1$.
  \textbf{B.} Contour plots of the variational posterior after 100 likelihood evaluations, with the noise-adjusted $a_\text{npro}$ acquisition function (left) and the newly proposed $a_\text{VIQR}$ (right). Red crosses indicate the centers of the variational mixture components, black dots are the training samples. \textbf{C.} ELBO as a function of likelihood evaluations. Shaded area is 95\% CI of the ELBO estimated via Bayesian quadrature. Dashed line is the true log marginal likelihood (LML).
}
  \label{fig:demo}
\end{figure}

In the rest of this section, we describe several new acquisition functions for \VBMC specifically designed to deal with noisy log-likelihood evaluations. Figure \ref{fig:demo} shows \VBMC at work in a toy noisy scenario (a `banana' 2D posterior), for two acquisition functions introduced in this section. 

\vspace{-0.5em}
\paragraph{Predictions with noisy evaluations} A useful quantity for this section is $s^2_{\gpdata \cup \x_\star}(\x)$, the predicted posterior GP variance at $\x$ if we make a function evaluation at $\x_\star$, with $y_\star$ distributed according to the posterior \emph{predictive} distribution (that is, inclusive of observation noise $\sigma_\text{obs}(\x_\star)$), given training data $\gpdata$. Conveniently, $s^2_{\gpdata \cup \x_\star}(\x)$ can be expressed in closed form as
\begin{equation} \label{eq:predictive_var}
s^2_{\gpdata \cup \x_\star}(\x) =  \; s_{\gpdata}^2(\x) - \frac{\covar_\gpdata^2(\x,\x_\star)}{\covar_\gpdata(\x_\star,\x_\star) + \sigma^2_\text{obs}(\x_\star)},
\end{equation}
where $C_{\gpdata}(\cdot,\cdot)$ denotes the GP posterior covariance (see \cite[Lemma 5.1]{jarvenpaa2020parallel}, and also \cite{lyu2018evaluating,jarvenpaa2019efficient}).

\vspace{-0.25em}
\subsection{Noisy prospective uncertainty sampling} 

The rationale behind $a_\text{pro}$ (Eq. \ref{eq:acqpro}) and similar heuristic `uncertainty sampling' acquisition functions \cite{gunter2014sampling,acerbi2019exploration} is to evaluate the log joint where the pointwise variance of the integrand in the expected log joint (as per Eq. \ref{eq:logjoint}, or variants thereof) is maximum.
For noiseless evaluations, this choice is equivalent to maximizing the \emph{variance reduction} of the integrand after an observation. 
Considering the GP posterior variance reduction, $\Delta s^2_{\gpdata}(\x) \equiv s^2_{\gpdata}(\x) - s^2_{\gpdata \cup \x}(\x)$, we see that, in the absence of observation noise, $s^2_{\gpdata \cup \x}(\x) = 0$ and $\Delta s^2(\x)_{\gpdata} = s^2_{\gpdata}(\x)$.
Thus, a natural generalization of uncertainty sampling to the noisy case is obtained by switching the GP posterior variance in Eq. \ref{eq:acqpro} to the GP posterior variance \emph{reduction}. Improving over the original uncertainty sampling, this generalization accounts for potential observation noise at the candidate location.

Following this reasoning, we generalize uncertainty sampling to noisy observations by defining the \emph{noise-adjusted prospective uncertainty sampling} acquisition function,
\begin{equation} \label{eq:acqnpro}
a_\text{npro}(\x) = \Delta s^2_{\gpdata}(\x) q_{\qparams}(\x) \exp\left( \overline{f}_{\gpdata}(\x)\right) = \left(\frac{s^2_{\gpdata}(\x)}{s^2_{\gpdata}(\x) + \sigman^2(\x)}\right) s^2_{\gpdata}(\x) q_{\qparams}(\x) \exp\left( \overline{f}_{\gpdata}(\x)\right),
\end{equation}
where we used Eq. \ref{eq:predictive_var} to calculate $s^2_{\gpdata \cup \x}(\x)$. Comparing Eq. \ref{eq:acqnpro} to Eq. \ref{eq:acqpro}, we see that $a_\text{npro}$ has an additional multiplicative term that accounts for the residual variance due to a potentially noisy observation.
As expected, it is easy to see that $a_\text{npro}(\x) \rightarrow a_\text{pro}(\x)$ for $\sigman(\x) \rightarrow 0$.

While $a_\text{npro}$ and other forms of uncertainty sampling operate pointwise on the posterior \emph{density}, we consider next \emph{global} (integrated) acquisition functions that account for non-local changes in the GP surrogate model when making a new observation, thus driven by uncertanty in posterior \emph{mass}.

\vspace{-0.25em}
\subsection{Expected information gain (\EIG)}

A principled information-theoretical approach suggests to sample points that maximize the \emph{expected information gain} (\EIG) about the integral of interest (Eq. \ref{eq:logjoint}). Following recent work on multi-source active-sampling Bayesian quadrature \cite{gessner2019active}, we can do so by choosing the next location $\x_\star$ that maximizes the \emph{mutual information} $I\left[\logjoint; y_\star\right]$ between the expected log joint $\logjoint$ and a new (unknown) observation $y_\star$. 
Since all involved quantities are jointly Gaussian, we obtain
\begin{equation} \label{eq:eig}
a_\text{EIG}(\x) = -\frac{1}{2} \log \left(1 - \rho^2(\x) \right), \qquad \text{with} \;
\rho(\x) \equiv \frac{\mathbb{E}_{\qparams}\left[ C_\gpdata(f(\cdot), f(\x)) \right]}{\sqrt{v_{\gpdata}(\x) \mathbb{V}_{f|\gpdata}[\logjoint]}},
\end{equation}
where $\rho(\cdot)$ is the \emph{scalar correlation} \cite{gessner2019active}, $v_{\gpdata}(\cdot)$ the GP posterior predictive variance (including observation noise), and $\mathbb{V}_{f|\gpdata}[\logjoint]$ the posterior variance of the expected log joint -- all given the current training set $\gpdata$.
The scalar correlation in Eq. \ref{eq:eig} has a closed-form solution thanks to Bayesian quadrature (see \supplement for derivations).

\vspace{-0.25em}
\subsection{Integrated median / variational interquantile range (\IMIQR / \VIQR)} 

J\"arvenp\"a\"a and colleagues \cite{jarvenpaa2020parallel} recently proposed the \emph{interquantile range} (\textsc{iqr}) as a robust estimate of the uncertainty of the unnormalized posterior, as opposed to the variance, 
and derived the \emph{integrated median interquantile range} (\IMIQR) acquisition function from Bayesian decision theory,
\begin{equation} \label{eq:imiqr}
a_\text{IMIQR}(\x) = - 2 \int_{\X} \exp\left({\overline{f}_{\gpdata}(\x^\prime)}\right) \sinh\left(u s_{\gpdata \cup \x}(\x^\prime) \right) d\x^\prime,
\end{equation}
where $u \equiv \Phi^{-1}(p_u)$, with $\Phi$ the standard normal \textsc{cdf} and $p_u \in (0.5,1)$ a chosen quantile (we use $p_u = 0.75$ as in \cite{jarvenpaa2020parallel}); $\sinh(z) = (\exp(z) -\exp(-z))/2$ for $z \in \mathbb{R}$ is the hyperbolic sine; and $s_{\gpdata \cup \x}(\x^\prime)$ denotes the predicted posterior standard deviation after observing the function at $\x^\prime$, 
as per Eq. \ref{eq:predictive_var}.
However, the integral in Eq. \ref{eq:imiqr} is intractable, and thus needs to be approximated at a significant computational cost (e.g., via \MCMC and importance sampling \cite{jarvenpaa2020parallel}). 

Instead, we note that the term $\exp\left(\overline{f}_{\gpdata}\right)$ in Eq. \ref{eq:imiqr} represents the joint distribution as modeled via the GP surrogate, which \VBMC further approximates with the variational posterior $q_{\qparams}$ (up to a normalization constant). Thus, 
we exploit the variational approximation of \VBMC to propose here the \emph{variational} (integrated median) \emph{interquantile range} (\VIQR) acquisition function,
\begin{equation} \label{eq:vimiqr}
a_\text{VIQR}(\x) = - 2\int_{\X} q_{\qparams}(\x^\prime) \sinh\left(u s_{\gpdata \cup \x}(\x^\prime) \right) d\x^\prime,
\end{equation}
where we replaced the surrogate posterior in Eq. \ref{eq:imiqr} with its corresponding variational posterior. Crucially, Eq. \ref{eq:vimiqr} can be approximated very cheaply via simple Monte Carlo by drawing $N_\text{viqr}$ samples from $q_{\qparams}$ (we use $N_\text{viqr} = 100$). In brief, $a_\text{VIQR}$ obtains a computational advantage over $a_\text{IMIQR}$ at the cost of adding a layer of approximation in the acquisition function ($q_{\qparams} \approx \exp\left(\overline{f}_{\gpdata}\right)$), but it otherwise follows from the same principles. Whether this approximation is effective in practice is an empirical question that we address in the next section.


\vspace{-0.25em}
\section{Experiments}

We tested different versions of \VBMC and other surrogate-based inference algorithms on a novel benchmark problem set consisting of a variety of computational models applied to real data (see Section \ref{sec:problems}).
For each problem, the goal of inference is to approximate the posterior distribution and the log marginal likelihood (LML) with a fixed budget of likelihood evaluations.

\vspace{-0.5em}
\paragraph{Algorithms} In this work, we focus on comparing new acquisition functions for \VBMC which support noisy likelihood evaluations, that is $a_\text{npro}$, $a_\text{EIG}$, $a_\text{IMIQR}$ and $a_\text{VIQR}$ as described in Section \ref{sec:noisy} (denoted as \vbmc plus, respectively, \npro, \eig, \imiqr or \viqr).
As a strong baseline for posterior estimation, we test a state-of-the-art technique for Bayesian inference via GP surrogates, which also uses $a_\text{IMIQR}$ \cite{jarvenpaa2020parallel} (\gpimiqr). \gpimiqr was recently shown to decisively outperform several other surrogate-based methods for posterior estimation in the presence of noisy likelihoods \cite{jarvenpaa2020parallel}.
For model evidence evaluation, to our knowledge no previous surrogate-based technique explicitly supports noisy evaluations. We test as a baseline \emph{warped sequential active Bayesian integration} (\wsabi \cite{gunter2014sampling}), a competitive method in a previous noiseless comparison \cite{acerbi2018variational}, adapted here for our benchmark (see \supplement). For each algorithm, we use the same default settings across problems.
We do not consider here non-surrogate based methods, such as Monte Carlo and importance sampling, which performed poorly with a limited budget of likelihood evaluations already in the noiseless case \cite{acerbi2018variational}.


\vspace{-0.5em}
\paragraph{Procedure}
For each problem, we allow a budget of 50$\times$($\nparams$+2) likelihood evaluations. For each algorithm, we performed 100 runs per problem with random starting points, and we evaluated performance with several metrics (see Section \ref{sec:results}). For each metric, we report as a function of likelihood evaluations the median and $95\%$ CI of the median calculated by bootstrap (see \supplement for a `worse-case' analysis of performance). For algorithms other than \VBMC, we only report metrics they were designed for (posterior estimation for \gpimiqr, model evidence for \wsabi).

\vspace{-0.5em}
\paragraph{Noisy log-likelihoods} For a given data set, model and parameter vector $\vtheta$, we obtain noisy evaluations of the log-likelihood through different methods, depending on the problem.  In the \emph{synthetic likelihood} (SL) approach, we run $N_\text{sim}$ simulations for each evaluation, and estimate the log-likelihood of summary statistics of the data under a multivariate normal assumption \cite{wood2010statistical,price2018bayesian,jarvenpaa2020parallel}. With \emph{inverse binomial sampling} (IBS), we obtain unbiased estimates of the log-likelihood of an entire data set by sampling from the model until we obtain a `hit' for each data point \cite{haldane1945method,van2020unbiased}; we repeat the process  $N_\text{rep}$ times and average the estimates for higher precision. Finally, for a few analyses we `emulate' noisy evaluations by adding i.i.d. Gaussian noise to deterministic log-likelihoods. Despite its simplicity, the `emulated noise' approach is statistically similar to IBS, as IBS estimates are unbiased, normally-distributed, and with near-constant variance across the parameter space \cite{van2020unbiased}.

\vspace{-0.25em}
\subsection{Benchmark problems}
\label{sec:problems}

The benchmark problem set consists of a common test simulation model (the Ricker model \cite{wood2010statistical}) and five models with real data from various branches of computational and cognitive neuroscience. Some models are applied to multiple datasets, for a total of nine inference problems with $3 \le \nparams \le 9$ parameters.
Each problem provides a target noisy log-likelihood, and for simplicity 
we assume a uniform prior over a bounded interval for each parameter.
For the purpose of this benchmark, we chose tractable models so that we could compute ground-truth posteriors and model evidence via extensive \MCMC sampling.
We now briefly describe each model; see \supplement for more details.

\vspace{-0.5em}
\paragraph{Ricker}

The Ricker model is a classic population model used in computational ecology \cite{wood2010statistical}. The population size $N_t$ evolves according to a discrete-time stochastic process $N_{t+1} = r N_t \exp\left(-N_t + \varepsilon_t \right)$, for $t = 1,\ldots,T$, with $\varepsilon_t \stackrel{\text{i.i.d.}}{\sim}  \mathcal{N}\left(0, \sigma_\varepsilon^2 \right)$ and $N_0 = 1$. At each time point, we have access to a noisy measurement $z_t$ of the population size $N_t$ with Poisson observation model $z_t \sim \text{Poisson}(\phi N_t)$. The model parameters are $\vtheta = (\log(r),\phi,\sigma_\varepsilon)$. We generate a dataset of observations $\z = (z_t)_{t=1}^T$ using the ``true'' parameter vector $\vtheta_\text{true} = (3.8,10,0.3)$ with $T = 50$, as in \cite{jarvenpaa2020parallel}.
We estimate the log-likelihood via the log-SL approach using the same 13 summary statistics as in \cite{wood2010statistical,gutmann2016bayesian,price2018bayesian,jarvenpaa2020parallel}, with $N_\text{sim} = 100$ simulations per evaluation, which yields $\sigman(\vtheta_\text{MAP}) \approx 1.3$, where $\vtheta_\text{MAP}$ is the maximum-a-posteriori (MAP) parameter estimate found via optimization. 

\vspace{-0.5em}
\paragraph{Attentional drift-diffusion model (aDDM)} The \emph{attentional drift-diffusion model} (aDDM) is a seminal model for value-based decision making between two items with ratings $r_\text{A}$ and $r_\text{B}$ \cite{krajbich2010visual}. 
At each time step $t$, the decision variable $\ddx_t$ is assumed to follow a stochastic diffusion process
\vspace{-0.25em}
\begin{equation}
\ddx_0 = 0, \qquad \ddx_{t+\delta t} = \ddx_{t} +  d \left(\att^{a_t} r_\text{A} - \att^{(1-a_t)} r_\text{B}\right) \delta t + \varepsilon_t, \qquad \varepsilon_t \stackrel{\text{i.i.d.}}{\sim}  \mathcal{N}\left(0, \sigma_\varepsilon^2 \delta t \right),
\end{equation} 
where $\varepsilon_t$ is the diffusion noise; $d$ is the drift rate; $\att\in [0,1]$ is the attentional bias factor; and $a_t = 1$ (resp., $a_t = 0$) if the subject is fixating item A (resp., item B) at time $t$. Diffusion continues until the decision variable hits the boundary $|\ddx_t| \ge 1$, which induces a choice (A for +1, B for -1). We include a lapse probability $\lambda$ of a random choice at a uniformly random time over the maximum trial duration, and set $\delta t = 0.1$ s. The model has parameters $\vtheta = (d, \att, \sigma_\varepsilon, \lambda)$.
We fit choices and reaction times of two subjects (S1 and S2) from \cite{krajbich2010visual} using IBS with $N_\text{rep} = 500$, which produces $\sigman(\vtheta_\text{MAP}) \approx 2.8$.

\vspace{-0.5em}
\paragraph{Bayesian timing} We consider a popular model of Bayesian time perception \cite{jazayeri2010temporal,acerbi2012internal}. In each trial of a sensorimotor timing task, human subjects had to reproduce the time interval $\tau$ between a click and a flash, with $\tau \sim \text{Uniform}$[0.6, 0.975] s \cite{acerbi2012internal}. We assume subjects had only access to a noisy sensory measurement $t_\text{s} \sim \mathcal{N}\left(\tau, w_\text{s}^2 \tau^2 \right)$, and their reproduced time $t_\text{m}$ was affected by motor noise, $t_\text{m} \sim \mathcal{N}\left(\tau_\star, w_m^2 \tau_\star^2 \right)$, where $w_\text{s}$ and $w_\text{m}$ are Weber's fractions. We assume subjects estimated $\tau_\star$ by combining their sensory likelihood with an approximate Gaussian prior over time intervals, $\normpdf{\tau}{\mu_\text{p}}{\sigma^2_\text{p}}$, and took the mean of the resulting Bayesian posterior. For each trial we also consider a probability $\lambda$ of a `lapse' (e.g., a misclick) producing a response $t_\text{m} \sim \text{Uniform}$[0, 2] s.  Model parameters are $\vtheta = (w_\text{s},w_\text{m},\mu_\text{p},\sigma_\text{p},\lambda)$.
We fit timing responses (discretized with $\delta t_\text{m} = 0.02$ s) of a representative subject from \cite{acerbi2012internal} using IBS with $N_\text{rep} = 500$, which yields $\sigman(\vtheta_\text{MAP}) \approx 2.2$.

\ignore{
\vspace{-0.5em}
\paragraph{g-and-k}

The g-and-k model is a common benchmark simulation model represented by a flexible probability distribution defined via its quantile function,
\begin{equation} \label{eq:gandk}
Q\left(\Phi^{-1}(p); \vtheta \right) = a + b \left(1 + c \frac{1 - \exp\left(-g \Phi^{-1}(p)\right)}{1 + \exp\left(-g \Phi^{-1}(p)\right)} \right) \left[1 + \left(\Phi^{-1}(p) \right)^2 \right]^k \Phi^{-1}(p),
\end{equation}
where $a,b,c,g$ and $k$ are parameters and $p \in [0,1]$ is a quantile. We fix $c = 0.8$ and consider the parameters $\vtheta = (a,b,g,k)$. We use the same dataset as \cite{price2018bayesian,jarvenpaa2020parallel}, generated with ``true'' parameter vector $\vtheta_\text{true} = (3,1,2,0.5)$, and for the log-SL estimation the same four summary statistics obtained by fitting a skew $t$-distribution to a set of samples generated from Eq. \ref{eq:gandk}.
Again, we use $N_\text{sim} = 100$, which produces fairly precise observations near the model area, with $\sigman(\vtheta_\text{MAP}) \approx 0.14$.
}

\vspace{-0.5em}
\paragraph{Multisensory causal inference (CI)}

\emph{Causal inference} (CI) in multisensory perception denotes the problem the brain faces when deciding whether distinct sensory cues come from the same source \cite{kording2007causal}. We model a visuo-vestibular CI experiment in which human subjects, sitting in a moving chair, were asked in each trial whether the direction of movement $s_\text{vest}$ matched the direction $s_\text{vis}$ of a looming visual field \cite{acerbi2018bayesian}. We assume subjects only have access to noisy sensory measurements $z_\text{vest} \sim \mathcal{N}\left(s_\text{vest}, \sigma^2_\text{vest} \right)$, $z_\text{vis} \sim \mathcal{N}\left(s_\text{vis}, \sigma^2_\text{vis}(c) \right)$, where $\sigma_\text{vest}$ is the vestibular noise and $\sigma_\text{vis}(c)$ is the visual noise, with $c \in \{c_\text{low}, c_\text{med}, c_\text{high}\}$  distinct levels of visual coherence adopted in the experiment.
We model subjects' responses with a heuristic `Fixed' rule that judges the source to be the same if $|z_\text{vis} - z_\text{vest}| < \kappa$, plus a probability $\lambda$ of giving a random response (\emph{lapse}) \cite{acerbi2018bayesian}. Model parameters are $\vtheta = (\sigma_\text{vest},\sigma_\text{vis}(c_\text{low}),\sigma_\text{vis}(c_\text{med}),\sigma_\text{vis}(c_\text{high}),\kappa,\lambda)$. We fit datasets from two subjects (S1 and S2) from \cite{acerbi2018bayesian} using IBS with $N_\text{rep} = 200$ repeats, which yields $\sigman(\vtheta_\text{MAP}) \approx 1.3$ for both datasets. 

\vspace{-0.5em}
\paragraph{Neuronal selectivity}

We consider a computational model of neuronal orientation selectivity in visual cortex \cite{goris2015origin} used in previous optimization and inference benchmarks \cite{acerbi2017practical,acerbi2018variational,acerbi2019exploration}. It is a linear-nonlinear-linear-nonlinear (LN-LN) cascade model which combines effects of filtering, suppression, and response nonlinearity whose output drives the firing rate of an inhomogeneous Poisson process (details in \cite{goris2015origin}). The restricted model has $\nparams = 7$ free parameters which determine features such as the neuron's preferred direction of motion and spatial frequency. We fit the neural recordings of one V1 and one V2 cell from \cite{goris2015origin}. 
For the purpose of this `noisy' benchmark, we compute the log-likelihood exactly and 
add i.i.d. Gaussian noise to each log-likelihood evaluation with $\sigman(\vtheta) = 2$.

\vspace{-0.5em}
\paragraph{Rodent 2AFC}

We consider a sensory-history-dependent model of rodent decision making in a two-alternative forced choice (2AFC) task. In each trial, rats had to discriminate the amplitudes $s_\text{L}$ and $s_\text{R}$ of auditory tones presented, respectively, left and right \cite{akrami2018posterior,roy2018efficient}. The rodent's choice probability is modeled as $P(\text{Left}) = \lambda/2 + (1-\lambda) / (1 + e^{-A})$ where $\lambda$ is a lapse probability and
\begin{equation}
A = w_{0} + w_{\text{c}} b_\text{c}^{(-1)} + w_{\overline{s}} \overline{s} + \sum_{t = 0}^2 \left(w_\text{L}^{(-t)} s_\text{L}^{(-t)} + w_\text{R}^{(-t)} s_\text{R}^{(-t)} \right),
\end{equation}
where $w_\text{L}^{(-t)}$ and $w_\text{R}^{(-t)}$ are coefficients of the $s_\text{L}$ and $s_\text{R}$ regressors, respectively, from $t$ trials back; $b_\text{c}^{(-1)}$ is the correct side on the previous trial ($\text{L} = +1$, $\text{R} = -1$), used to capture the win-stay/lose-switch strategy; $\overline{s}$ is a long-term history regressor (an exponentially-weighted running mean of past stimuli with time constant $\tau$); and $w_0$ is the bias. 
This choice of regressors best described rodents' behavior in the task \cite{akrami2018posterior}.
We fix $\lambda = 0.02$ and $\tau = 20$ trials, thus leaving $\nparams = 9$ free parameters $\vtheta = (w_0,w_{\text{c}},w_{\overline{s}},\bm{w}_\text{L}^{(0,-1,-2)},\bm{w}_\text{R}^{(0,-1,-2)})$. 
We fit $10^4$ trials from a representative subject dataset \cite{roy2018efficient} using IBS with $N_\text{rep} = 500$, which produces $\sigman(\vtheta_\text{MAP}) \approx 3.18$.

\vspace{-0.25em}
\subsection{Results}
\label{sec:results}

\begin{figure}[htb]
  \includegraphics[width=\linewidth]{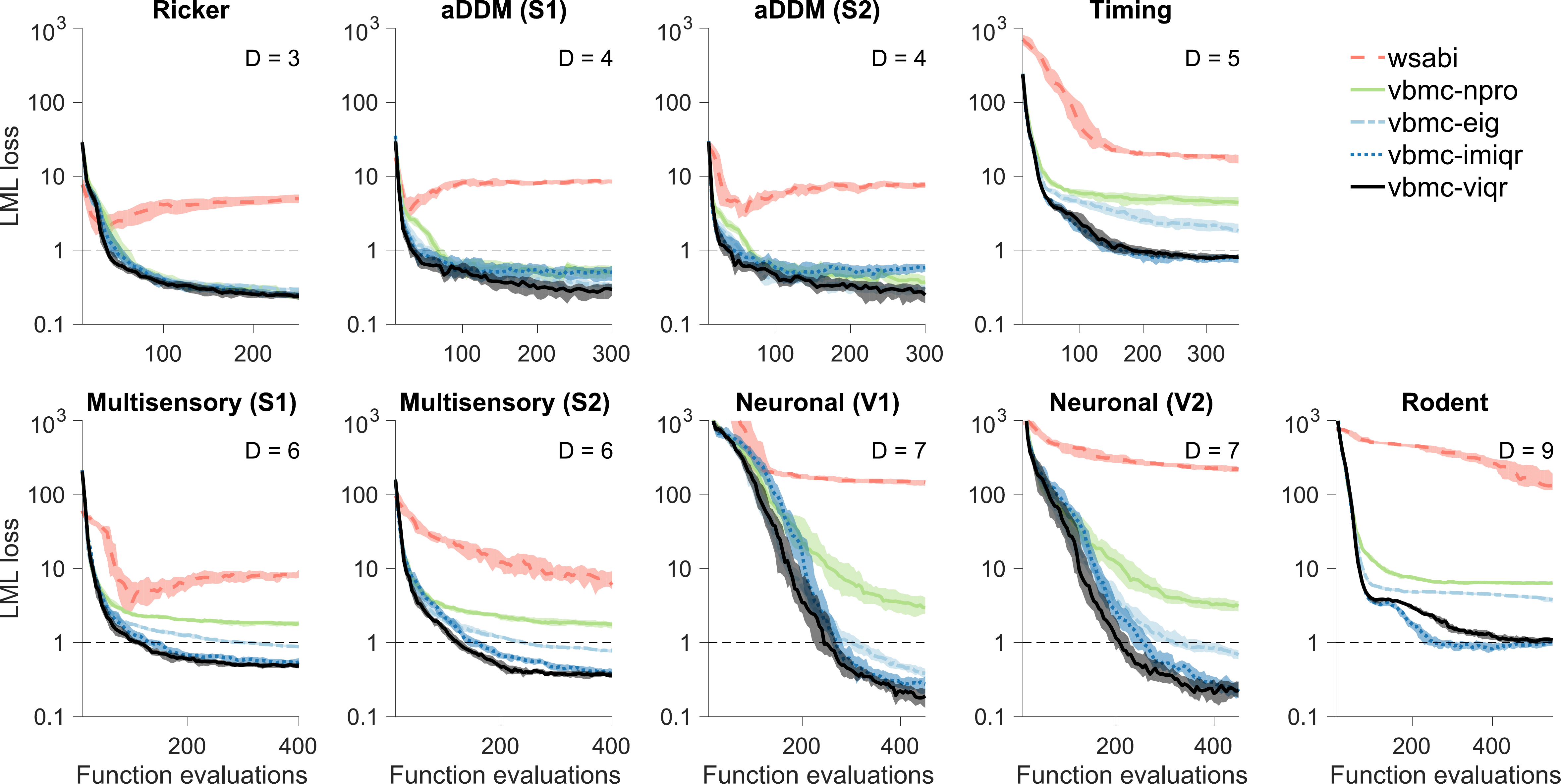}  
\vspace{-1.25em}
\caption{{\bf Model evidence loss.} Median absolute error of the log marginal likelihood (LML) estimate with respect to ground truth, as a function of number of likelihood evaluations, on different problems. A desirable error is below 1 (dashed line). Shaded areas are 95\% CI of the median across 100 runs.}
  \label{fig:lml}
\end{figure}

To assess the model evidence approximation, Fig. \ref{fig:lml} shows the absolute difference between true and estimated log marginal likelihood (`LML loss'), using the ELBO as a proxy for \VBMC. Differences in LML of 10+ points are often considered `decisive evidence' in a model comparison  \cite{kass1995bayes}, while differences $\ll 1$ are negligible; so for practical usability of a method we aim for a LML loss $< 1$. 

\begin{figure}[htb]
  \includegraphics[width=\linewidth]{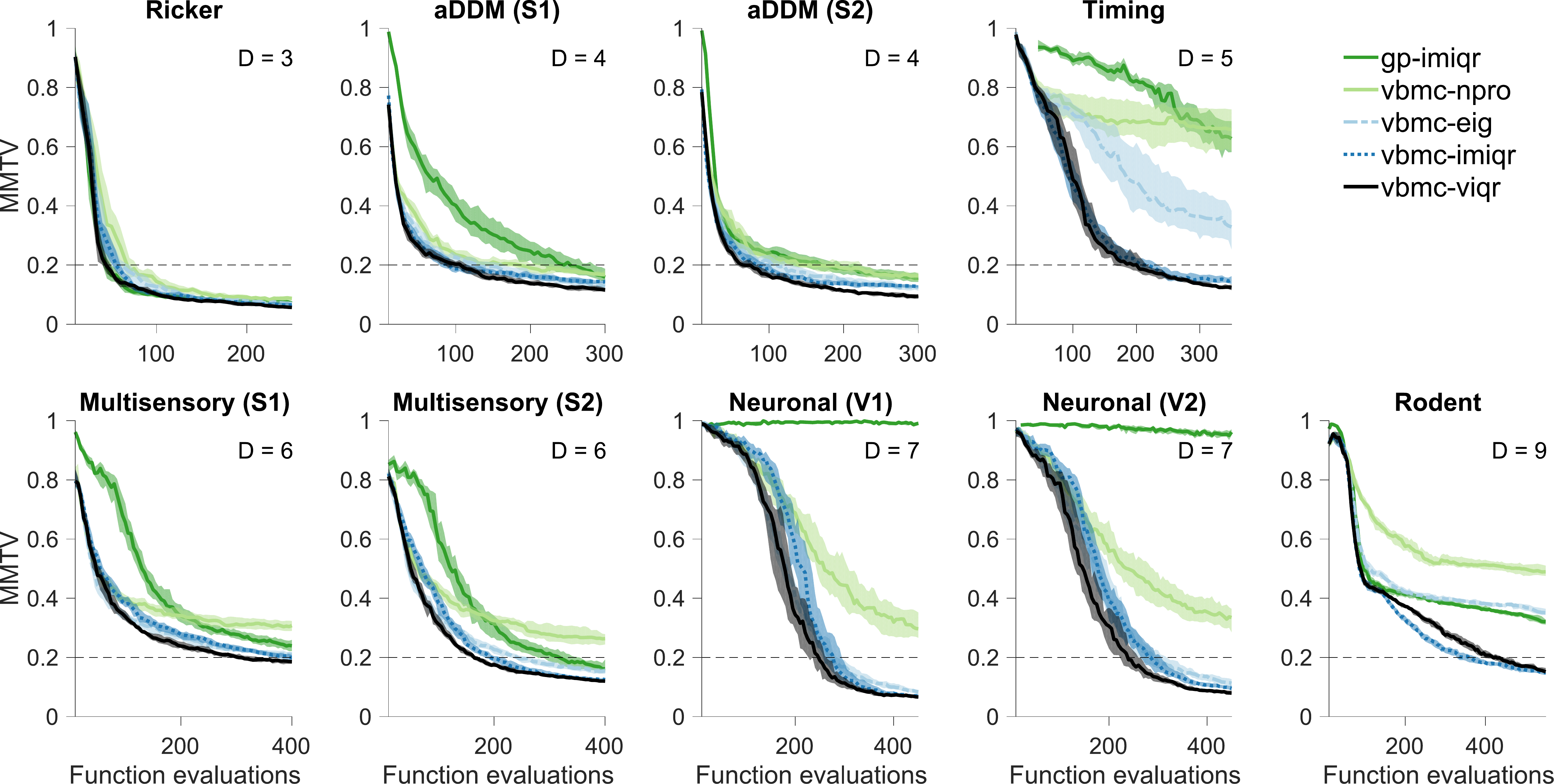}  
\vspace{-1.25em}
\caption{{\bf Posterior estimation loss (MMTV).} Median mean marginal total variation distance (MMTV) between the algorithm's posterior and ground truth, as a function of number of likelihood evaluations. A desirable target (dashed line) is less than 0.2, corresponding to more than 80\% overlap between true and approximate posterior marginals (on average across model parameters). 
}
  \label{fig:mmtv}
\end{figure}

As a measure of loss to judge the quality of the posterior approximation, Fig. \ref{fig:mmtv} shows the mean marginal total variation distance (MMTV) between approximate posterior and ground truth.
Given two pdfs $p$ and $q$, we define $\text{MMTV}(p,q) = \frac{1}{2\nparams} \sum_{i=1}^\nparams \int \left| p_i(x_i) - q_i(x_i) \right| d x_i $, where $p_i$ and $q_i$ denote the marginal densities along the $i$-th dimension. Since the MMTV only looks at differences in the marginals, we also examined the ``Gaussianized'' symmetrized Kullback-Leibler divergence (gsKL), a metric sensitive to differences in mean and covariance \cite{acerbi2018variational}. We found that MMTV and gsKL follow qualitatively similar trends, so we show the latter in the \supplement. 

First, our results confirm that, in the presence of noisy log-likelihoods, methods that use `global' acquisition functions largely outperform methods that use pointwise estimates of uncertainty, as noted in \cite{jarvenpaa2020parallel}. In particular, `uncertainty sampling' acquisition functions are unusable with \VBMC in the presence of noise, exemplified here by the poor performance of \vbmc-\npro (see also \supplement for further tests). \wsabi shows the worst performance here due to a GP representation (the square root transform) which interacts badly with noise on the log-likelihood.
Previous state-of-the art method \gpimiqr performs well with a simple synthetic problem (Ricker), but fails on complex scenarios such as Rodent 2AFC, Neuronal selectivity, or Bayesian timing, likely due to excessive exploration (see \supplement).
\vbmc-\eig performs reasonably well on most problems, but also struggles on Rodent 2AFC and Bayesian timing. 
Overall, \vbmc-\imiqr and \vbmc-\viqr systematically show the best and most robust performance, with \vbmc-\viqr marginally better on most problems, except Rodent 2AFC. Both achieve good approximations of the model evidence and of the true posteriors within the limited budget (see \supplement for comparisons with ground-truth posteriors).

Table \ref{tab:overhead} compares the average algorithmic overhead of methods based on $a_\text{IMIQR}$ and $a_\text{VIQR}$, showing the computational advantage of the variational approach of \vbmc-\viqr. 

\begin{table}[htb]
\caption{Average algorithmic overhead per likelihood evaluation (in seconds) over a full run, assessed on a single-core reference machine (mean $\pm$ 1 SD across 100 runs).}
  \label{tab:overhead}
  \centering
  \begin{tabular}{lcccccc}
    \toprule
    & \multicolumn{6}{c}{Model}                   \\
    \cmidrule(r){2-7}
    Algorithm     & Ricker   & aDDM & Timing & Multisensory & Neuronal & Rodent \\
    \midrule
\vbmc-\viqr & $\bm{1.5 \pm 0.1}$ & $\bm{1.5 \pm 0.1}$ & $\bm{1.8 \pm 0.2}$ & $\bm{2.0 \pm 0.2}$ & $\bm{2.8 \pm 0.8}$ & $\bm{2.6 \pm 0.2}$ \\
\vbmc-\imiqr & $5.5 \pm 0.5$ & $5.1 \pm 0.3$ & $5.8 \pm 0.6$ & $5.6 \pm 0.3$ & $6.5 \pm 1.3$ & $5.6 \pm 0.4$ \\
\gpimiqr & $15.6 \pm 0.9$ & $16.0 \pm 1.7$ & $17.1 \pm 1.2$ & $26.3 \pm 1.8$ & $29.6 \pm 2.8$ & $40.1 \pm 2.1$ \\
    \bottomrule
  \end{tabular}
\end{table}

\begin{figure}[htb]
  \includegraphics[width=\linewidth]{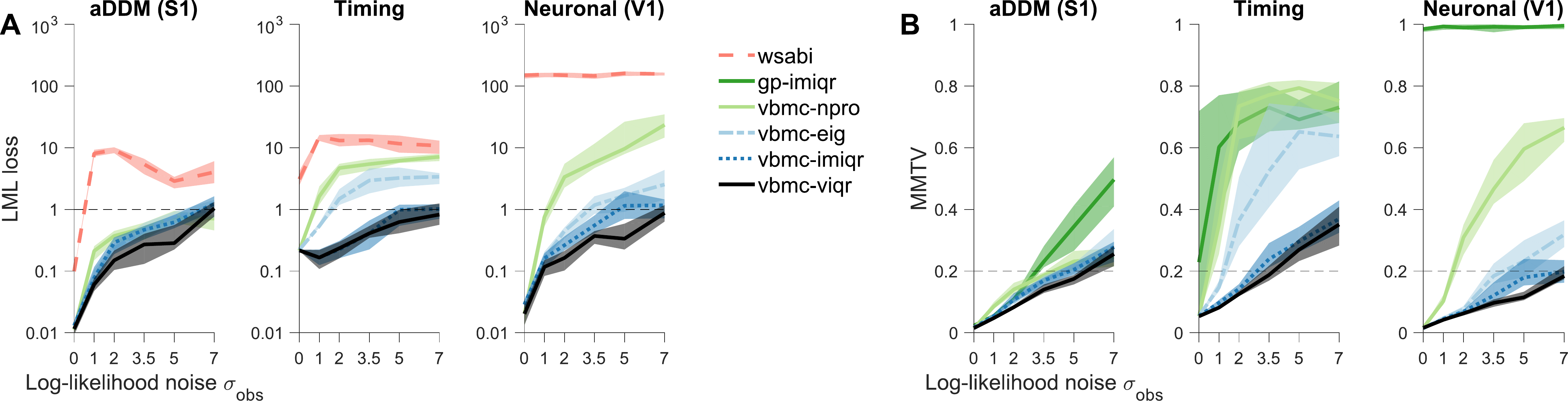}  
\caption{{\bf Noise sensitivity.} Final performance metrics of all algorithms with respect to ground truth, as a function of log-likelihood observation noise $\sigman$, for different problems. For all metrics, we plot the median after 50$\times$($\nparams$+2) log-likelihood evaluations, and shaded areas are 95 \% CI of the median across 100 runs. \textbf{A.} Absolute error of the log marginal likelihood (LML) estimate. \textbf{B.} Mean marginal total variation distance (MMTV).}
  \label{fig:noise}
\end{figure}

Then, we looked at how robust different methods are to different degrees of log-likelihood noise. We considered three benchmark problems for which we could easily compute the log-likelihood exactly. For each problem, we emulated different levels of noise by adding Gaussian observation noise to exact log-likelihood evaluations, with $\sigman \in [0, 7]$ (see Fig. \ref{fig:noise}).
Most algorithms only perform well with no or very little noise, whereas the performance of \vbmc-\viqr (and, similarly, \vbmc-\imiqr) degrades gradually with increasing noise. For these two algorithms, acceptable results can be reached for $\sigman$ as high as $\approx 7$, although for best results even with hard problems we would recommend $\sigman \lesssim 3$. We see that the Neuronal problem is particularly hard, with both \wsabi and \gpimiqr failing to converge altogether even in the absence of noise.


Lastly, we tested how robust \vbmc-\viqr is to imprecise estimates of the observation noise, $\widehat{\sigma}_\text{obs}(\x)$. We reran \vbmc-\viqr on the three problems of Fig. \ref{fig:noise} while drawing $\widehat{\sigma}_\text{obs} \sim \text{Lognormal}\left(\ln \sigma_\text{obs}, \sigma_\sigma^2 \right)$ for increasing values of noise-of-estimating-noise, $\sigma_\sigma \ge 0$. We found that at worst the performance of \vbmc degrades only by $\sim 25\%$ with $\sigma_\sigma$ up to $0.4$ (i.e., $\widehat{\sigma}_\text{obs}$ roughly between $0.5-2.2$ times the true value); showing that \vbmc is robust to imprecise noise estimates (see Supplement for details).




\vspace{-0.25em}
\section{Conclusions}

\vspace{-0.25em}
In this paper, we addressed the problem of approximate Bayesian inference with only a limited budget of noisy log-likelihood evaluations. For this purpose, we extended the \VBMC framework to work in the presence of noise by testing several new acquisition functions and by introducing variational whitening for a more accurate posterior approximation. We showed that with these new features \VBMC achieves state-of-the-art inference performance on a novel challenging benchmark that uses a variety of models and real data sets from computational and cognitive neuroscience, covering areas such as neuronal modeling, human and rodent psychophysics, and value-based decision-making. 

Our benchmark also revealed that common synthetic test problems, such as the Ricker and g-and-k models (see \supplement for the latter), may be too simple for surrogate-based methods, as good performance on these problems (e.g., \gpimiqr) may not generalize to real models and datasets.

In conclusion, our extensive analyses show that \VBMC with the $a_\text{VIQR}$ acquisition function is very effective for approximate Bayesian inference with noisy log-likelihoods, with up to $\sigman \approx 3$, and models up to $\nparams \lesssim 10$ and whose evaluation take about a few seconds or more. Future work should focus on improving the flexibility of the GP representation, scaling the method to higher dimensions, and investigating theoretical guarantees for the \VBMC algorithm.




\clearpage
\section*{Broader Impact}

We believe this work has the potential to lead to net-positive improvements in the research community and more broadly in society at large.
First, this paper makes Bayesian inference accessible to non-cheap models with noisy log-likelihoods, allowing more researchers to express uncertainty about their models and model parameters of interest in a principled way; with all the advantages of proper uncertainty quantification \cite{gelman2013bayesian}.
Second, with the energy consumption of computing facilities growing incessantly every hour, it is our duty towards the environment to look for ways to reduce the carbon footprint of our algorithms \cite{strubell2019energy}. In particular, traditional methods for approximate Bayesian inference can be extremely sample-inefficient. 
The `smart' sample-efficiency of \VBMC can save a considerable amount of resources when model evaluations are computationally expensive.

Failures of \VBMC can return largely incorrect posteriors and values of the model evidence, which if taken at face value could lead to wrong conclusions. This failure mode is not unique to \VBMC, but a common problem of all approximate inference techniques (e.g., \MCMC or variational inference \cite{gelman2013bayesian,yao2018yes}). \VBMC returns uncertainty on its estimate and comes with a set of diagnostic functions which can help identify issues. Still, we recommend the user to follow standard good practices for validation of results, such as posterior predictive checks, or comparing results from different runs.

Finally, in terms of ethical aspects, our method -- like any general, black-box inference technique -- will reflect (or amplify) the explicit and implicit biases present in the models and in the data, especially with insufficient data \cite{chen2018my}. Thus, we encourage researchers in potentially sensitive domains to explicitly think about ethical issues and consequences of the models and data they are using.

\begin{ack}
We thank Ian Krajbich for sharing data for the aDDM model; Robbe Goris for sharing data and code for the neuronal model; Marko J\"arvenp\"a\"a and Alexandra Gessner for useful discussions about their respective work; Nisheet Patel for helpful comments on an earlier version of this manuscript; and the anonymous reviewers for constructive remarks.
This work has utilized the NYU IT High Performance Computing resources and services.
This work was partially supported by the Academy of Finland Flagship programme: Finnish Center for Artificial Intelligence (FCAI).
\end{ack}


\medskip
{
\small

}

\clearpage

\appendix
\part*{Supplementary Material}

\etocdepthtag.toc{mtappendix}
\etocsettagdepth{mtchapter}{none}
\etocsettagdepth{mtappendix}{subsubsection}

\setcounter{footnote}{0}
\setcounter{figure}{0}
\setcounter{table}{0}
\setcounter{equation}{0}
\renewcommand{\theequation}{S\arabic{equation}}
\renewcommand{\thetable}{S\arabic{table}}
\renewcommand{\thefigure}{S\arabic{figure}}

In this Supplement we include a number of derivations, implementation details, and additional results omitted from the main text.

Code used to generate the results and figures in the paper is available at \url{https://github.com/lacerbi/infbench}. 
The \VBMC algorithm with added support for noisy models is available at  \url{https://github.com/lacerbi/vbmc}.

\tableofcontents

\section{Background information}
\label{sec:background}

For ease of reference, in this Section we recap the three key theoretical ingredients used to build the Variational Bayesian Monte Carlo (\VBMC) framework, that is variational inference, Gaussian processes and adaptive Bayesian quadrature. The material presented here is largely based and expands on the ``theoretical background'' section of \cite{acerbi2018variational}.

\subsection{Variational inference}
\label{sec:variationalinference}

Let $\x \in \X \subseteq \mathbb{R}^\nparams$ be a parameter vector of a model of interest, and $\data$ a dataset. Variational inference is an approximate inference framework in which an intractable posterior $\post$ is approximated by a simpler distribution $\qx \equiv \qpx$ that belongs to a parametric family indexed by parameter vector $\qparams$, such as a multivariate normal or a mixture of Gaussians \cite{jordan1999introduction,bishop2006pattern}. 
Thus, the goal of variational inference is to find $\qparams$ for which the variational posterior $\qp$ is ``closest'' in approximation to the true posterior, according to some measure of discrepancy. 

In variational Bayes, the discrepancy between approximate and true posterior is quantified by the Kullback-Leibler (KL) divergence,
\begin{equation} \label{eq:KL}
D_{\text{KL}}\left[\qpx || \post\right] = \mathbb{E}_{\qparams} \left[\log \frac{q_{\qparams}(\x)}{\post} \right],
\end{equation}
where we adopted the compact notation $\mathbb{E}_{\qparams} \equiv \mathbb{E}_{q_{\qparams}}$. Crucially, $D_{\text{KL}}(q || p) \ge 0$ and the equality is achieved if and only if $q \equiv p$. $D_{\text{KL}}$ is not symmetric, and the specific choice of using $D_{\text{KL}}\left[q || p\right]$ (\emph{reverse} $D_{\text{KL}}$) as opposed to $D_{\text{KL}}\left[p || q\right]$ (\emph{forward} $D_{\text{KL}}$) is a key feature of the variational framework.

The variational approach casts Bayesian inference as an optimization problem, which consists of finding the variational parameter vector $\qparams$ that minimizes Eq. \ref{eq:KL}. 
We can rewrite Eq. \ref{eq:KL} as
\begin{equation} \label{eq:refactoring}
\log \ev = D_{\text{KL}}\left[\qpx || \post\right] + \mathcal{F}[q_{\qparams}],
\end{equation}
where on the left-hand side we have the \emph{model evidence}, and on the right-hand side the KL divergence plus the \emph{negative free energy}, defined as
\begin{equation} \label{eq:supp_elbo}
\mathcal{F}\left[ q_{\qparams} \right] =  \mathbb{E}_{\qparams} \left[\log \frac{\like p(\x)}{q_{\qparams}(\x)} \right] = \mathbb{E}_{\qparams} \left[f(\x) \right] + \mathcal{H}[q_{\qparams}(\x)],
\end{equation}
with $f(\x) \equiv \log \like p(\x) = \log p(\data, \x)$ the log joint probability, and $\mathcal{H}[q]$ the entropy of $q$.
Now, since as mentioned above the KL divergence is a non-negative quantity, 
from Eq. \ref{eq:refactoring} we have $\mathcal{F}[q] \le \log \ev$, with equality holding if $\qx \equiv \post$. For this reason, Eq. \ref{eq:supp_elbo} is known as the \emph{evidence lower bound} (ELBO), so called because it is a lower bound to the log marginal likelihood or model evidence.
Importantly, maximization of the variational objective, Eq. \ref{eq:supp_elbo}, is equivalent to minimization of the KL divergence, and produces both an approximation of the posterior $\qp$ and the ELBO, which can be used as a metric for model selection.

Classically, $q$ is chosen to belong to a family (e.g., a factorized posterior, or mean field) such that both the expected log joint in Eq. \ref{eq:supp_elbo} and the entropy afford analytical solutions, which are then used to yield closed-form equations for a coordinate ascent algorithm. In the \VBMC framework, instead, $f(\x)$ is assumed to be a potentially expensive black-box function, which prevents a direct computation of Eq. \ref{eq:supp_elbo} analytically or via simple numerical integration.

\subsection{Gaussian processes}

Gaussian processes (GPs) are a flexible class of statistical models for specifying prior distributions over unknown functions $f : \X \subseteq \mathbb{R}^{\nparams} \rightarrow \mathbb{R}$ \cite{rasmussen2006gaussian}. 
GPs are defined by a mean function $m: \X \rightarrow \mathbb{R}$  and a positive definite covariance, or kernel function $\kappa: \X \times \X \rightarrow \mathbb{R}$. \vbmc uses the common squared exponential (rescaled Gaussian) kernel,
\begin{equation} \label{eq:cov}
\kappa(\x,\x^\prime) = \sigma_f^2 \Lambda  \normpdf{\x}{\x^\prime}{\bm{\Sigma}_\ell} \qquad \text{with} \; \bm{\Sigma}_\ell = \text{diag}\left[{\ell^{(1)}}^2, \ldots, {\ell^{(\nparams)}}^2\right], 
\end{equation}
where $\sigma_f$ is the output length scale, $\bm{\ell}$ is the vector of input length scales, and $\Lambda \equiv \left(2\pi\right)^{\frac{\nparams}{2}} \prod_{i=1}^\nparams \ell^{(i)}$ is equal to the normalization factor of the Gaussian (this notation makes it easy to apply Gaussian identities used in Bayesian quadrature).
As a mean function, \vbmc uses a \emph{negative quadratic} function to ensure well-posedness of the variational formulation, and defined as \cite{acerbi2018variational,acerbi2019exploration}
\begin{equation}
m(\x) \equiv m_0 - \frac{1}{2} \sum_{i=1}^{\nparams} \frac{\left(\theta^{(i)} - \theta_\text{m}^{(i)}\right)^2}{{\omega^{(i)}}^2},
\end{equation}
where $m_0$ denotes the maximum, $\x_\text{m}$ is the location, and $\bm{\omega}$ is a vector of length scales.
Finally, GPs are also characterized by a likelihood or observation noise model, which is assumed here to be Gaussian with known variance $\sigman^2(\x)$ for each point in the training set (in the original formulation of \vbmc, observation noise is assumed to be a small positive constant).

Conditioned on training inputs $\xx = \left\{\x_1, \ldots,\x_N \right\}$, observed function values $\y = f(\xx)$ and observation noise $\sigman^2(\xx)$, the posterior GP mean and covariance are available in closed form \cite{rasmussen2006gaussian},
\begin{equation} \label{eq:postgp}
\begin{split}
\overline{f}_{\gpdata}(\x) \equiv \mathbb{E}\left[f(\x) | \gpdata, \vgp \right] = & \kappa(\x,\xx) \left[\kappa(\xx,\xx) + \bm{\Sigma}_\text{obs}(\xx)\right]^{-1} (\y - m(\xx)) + m(\x) \\
\covar_\gpdata(\x, \x^\prime) \equiv \text{Cov}\left[f(\x), f(\x^\prime) | \gpdata, \vgp\right] = & \; \kappa(\x,\x^\prime) - \kappa(\x,\xx) \left[\kappa(\xx,\xx) + \bm{\Sigma}_\text{obs}(\xx)\right]^{-1} \kappa(\xx,\x^\prime),
\end{split}
\end{equation}
where $\gpdata = \left\{\xx, \y, \bm{\sigma}_\text{obs} \right\}$ is the set of training function data for the GP; $\vgp$ is a hyperparameter vector for the GP mean, covariance, and likelihood; and $\bm{\Sigma}_\text{obs}(\xx) \equiv \text{diag}\left[\sigman^2(\x_1), \ldots, \sigman^2(\x_N)\right]$ is the observation noise (diagonal) matrix.

\subsection{Adaptive Bayesian quadrature}
\label{sec:bq}

Bayesian quadrature, also known as cubature when dealing with multi-dimensional integrals, is a technique to obtain Bayesian estimates of intractable integrals of the form \cite{ohagan1991bayes,ghahramani2003bayesian}
\begin{equation} \label{eq:integral}
Z = \int_{\X} f(\x) \pi(\x) d\x,
\end{equation}
where $f$ is a function of interest and $\pi$ a known probability distribution. For the purpose of \VBMC, we consider the domain of integration $\mathcal{X} = \mathbb{R}^\nparams$. 
When a GP prior is specified for $f$, since integration is a linear operator, the integral $Z$ is also a Gaussian random variable whose posterior mean and variance are \cite{ghahramani2003bayesian}
\begin{equation} \label{eq:bq}
\mathbb{E}_{f | \gpdata}[Z] = \int \overline{f}_\gpdata(\x) \pi(\x) d\x, \qquad
\mathbb{V}_{f | \gpdata}[Z] = \int \int \covar_\gpdata(\x, \x^\prime) \pi(\x) \pi(\x^\prime) d\x d\x^\prime.
\end{equation}
Importantly, if $f$ has a Gaussian kernel and $\pi$ is a Gaussian or mixture of Gaussians (among other functional forms), the integrals in Eq. \ref{eq:bq} have closed-form solutions. 

\vspace{-0.5em}
\paragraph{Active sampling}


The point $\x_\star \in \X$ to evaluate next to improve our estimate of the integral (Eq. \ref{eq:integral}) is chosen via a proxy optimization of a given \emph{acquisition function} $a: \X \rightarrow \mathbb{R}$, that is $\x_\star = \text{argmax}_{\x} a(\x)$. Previously introduced acquisition functions for Bayesian quadrature include the \emph{expected entropy}, which minimizes the expected entropy of the integral after adding $\x_\star$ to the training set \cite{osborne2012active}, and a family of strategies under the name of \emph{uncertainty sampling}, whose goal is generally to find the point with maximal (pointwise) variance of the \emph{integrand} at $\x_\star$ \cite{gunter2014sampling}. The standard acquisition function for \vbmc is \emph{prospective uncertainty sampling} (see main text and \cite{acerbi2018variational,acerbi2019exploration}).
Recent work proved convergence guarantees for active-sampling Bayesian quadrature under a broad class of acquisition functions which includes various forms of uncertainty sampling \cite{kanagawa2019convergence}.

\section{Algorithmic details}

We report here implementation details of new or improved features of the \VBMC algorithm omitted from the main text.

\subsection{Modified \vbmc features}
\label{sec:features}

In this section, we describe minor changes to the basic \vbmc framework.
For implementation details of the algorithm which have remained unchanged, we refer the reader to the main text and Supplement of the original \VBMC paper \cite{acerbi2018variational}.

\paragraph{Reliability index}

In \vbmc, the \emph{reliability index} $r(t)$ is a metric computed at the end of each iteration $t$ and determines, among other things, the termination condition \cite{acerbi2018variational}. We recall that $r(t)$ is computed as the arithmetic mean of three reliability features: 
\begin{enumerate}
\item The absolute change in mean ELBO from the previous iteration: $r_1(t) = \left|\mathbb{E}\left[\text{ELBO}(t)\right] - \mathbb{E}\left[\text{ELBO}(t-1)\right]\right|/\Delta_\text{SD}$.
\item The uncertainty of the current ELBO: $r_2(t) = \sqrt{\mathbb{V}\left[\text{ELBO}(t)\right]} / \Delta_\text{SD}$.
\item The change in `Gaussianized' symmetrized KL divergence (see Eq. \ref{eq:gsKL}) between the current and previous-iteration variational posterior $q_t \equiv q_{\qparams_t}(\x)$: $r(t) = \text{gsKL}(q_t || q_{t-1})/\Delta_\text{KL}$.
\end{enumerate}
The parameters $\Delta_\text{SD}$ and $\Delta_\text{KL}$ are tolerance hyperparameters, chosen such that $r_j \lesssim 1$, with $j = 1,2,3$, for features that are deemed indicative of a good solution. We set $\Delta_\text{KL} = 0.01 \cdot\sqrt{\nparams}$ as in the original \vbmc paper.
To account for noisy observations, we set $\Delta_\text{SD}$ in the current iteration equal to the geometric mean between the baseline $\Delta_\text{SD}^\text{base} = 0.1$ (from the original \vbmc paper) and the GP noise in the high-posterior density region, $\sigman^\text{hpd}$, and constrain it to be in the $[0.1, 1]$ range. That is,
\begin{equation}
\Delta_\text{SD} = \min \left[1, \max\left[0.1, \sqrt{\Delta_\text{SD}^\text{base} \cdot \sigman^\text{hpd}}\right]\right],
\end{equation}
where $\sigman^\text{hpd}$ is computed as the median observation noise at the top 20\% points in terms of log-posterior value in the GP training set.

\paragraph{Regularization of acquisition functions}
\label{sec:regularization}

In \VBMC, active sampling is performed by maximizing a chosen acquisition function $a : \X \subseteq \mathbb{R}^\nparams \rightarrow [0,\infty)$, where $\X$ is the support of the target density (see Section \ref{sec:supp_acqfuns}).
In practice, in \VBMC we maximize a \emph{regularized} acquisition function
\begin{equation} \label{eq:acqreg}
a^\text{reg}(\x; a) \equiv a(\x) b_\text{var}(\x) b_\text{bnd}(x)
\end{equation}
where $b_\text{var}(\x)$ is a GP variance regularization term introduced in \cite{acerbi2018variational},
\begin{equation} \label{eq:regvar}
b_\text{var}(\x) = \exp \left\{ - \left(\frac{V^\text{reg}}{V_{\gpdata}(\x)} - 1 \right) \left|\left[V_{\gpdata}(\x) < V^\text{reg} \right]\right| \right\}
\end{equation}
where $V_{\gpdata}(\x)$ is the posterior latent variance of the GP, $V^\text{reg}$ a regularization parameter (we use $V^\text{reg} = 10^{-4}$), and we denote with $|[\cdot]|$ \emph{Iverson's bracket} \cite{knuth1992two}, which takes value 1 if the expression inside the bracket is true, 0 otherwise. Eq. \ref{eq:regvar} penalizes the selection of points too close to an existing input, which might produce numerical issues.

The $b_\text{bnd}$ term is a new term that we added in this work to discard points too close to the parameter bounds, which would map to very large positive or negative values in the unbounded inference space,
\begin{equation} \label{eq:regbnd}
b_\text{bnd}(\x) = \left\{ \begin{array}{cl} 1 & \text{ if } \tilde{\theta}^{(i)} \ge \text{\texttt{LB}}_\varepsilon^{(i)} \wedge \tilde{\theta}^{(i)} \le \text{\texttt{UB}}_\varepsilon^{(i)}, \text{ for all } 1 \le i \le \nparams \\ 0 & \text{otherwise} \end{array} \right. 
\end{equation}
where $\tilde{\vtheta}(\vtheta)$ is the parameter vector remapped to the original space, and $\text{\texttt{LB}}_\varepsilon^{(i)} \equiv \text{\texttt{LB}}^{(i)} + \varepsilon (\text{\texttt{UB}}^{(i)} - \text{\texttt{LB}}^{(i)})$, $\text{\texttt{UB}}_\varepsilon^{(i)} \equiv \text{\texttt{UB}}^{(i)} - \varepsilon (\text{\texttt{UB}}^{(i)} - \text{\texttt{LB}}^{(i)})$, with $\varepsilon = 10^{-5}$.

\paragraph{GP hyperparameters and priors}

The GP model in \VBMC has $3\nparams+3$ hyperparameters, $\vgp = (\bm{\ell}, \sigma_f, \overline{\sigma}_\text{obs}, m_0, \x_\text{m}, \bm{\omega})$. All scale hyperparameters, that is $\left\{\bm{\ell}, \sigma_f, \overline{\sigma}_\text{obs}, \bm{\omega}\right\}$, are defined in log space. 
Each hyperparameter has an independent prior, either bounded uniform or a truncated Student's $t$ distribution with mean $\mu$, scale $\sigma$, and $\nu = 3$ degrees of freedom.
 GP hyperparameters and their priors are reported in Table \ref{tab:gphyp}.

\begin{table}[ht]
\begin{tabular}{clcccc}
Hyperparameter & Description & Prior mean $\mu$ & Prior scale $\sigma$ \\
\hline
$\log \ell^{(i)}$ & Input length scale & $\log \left[\sqrt{\frac{\nparams}{6}} L^{(i)}\right]$ & $\log \sqrt{10^3}$ 
\\
$\log \sigma_f$ & Output scale & Uniform & --- \\
$\log \overline{\sigma}_\text{obs}$ & Base observation noise & $\log \sqrt{10^{-5}}$ & 0.5 \\
$m_0$ & Mean function maximum & Uniform & --- \\
$x_\text{m}^{(i)}$ & Mean function location & Uniform & --- \\
$\log \omega^{(i)}$ & Mean function scale & Uniform & --- \\
\hline
\end{tabular}
  \centering
  \vspace{.5em}
\caption{GP hyperparameters and their priors. See text for more information.}
\label{tab:gphyp}
\end{table}

In Table \ref{tab:gphyp}, $\bm{L}$ denotes the vector of plausible ranges along each coordinate dimension, with $L^{(i)} \equiv \text{\texttt{PUB}}^{(i)} - \text{\texttt{PLB}}^{(i)}$. The base observation noise $\overline{\sigma}^2_\text{obs}$ is a constant added to the input-dependent observation noise $\sigman^2(\vtheta)$.
Note that we have modified the GP hyperparameter priors with respect to the original \vbmc paper, and these are now the default settings for both noisy and noiseless inference. In particular, we removed dependence of the priors from the GP training set (the `empirical Bayes' approach previously used), as it was found to occasionally generate unstable behavior.

\paragraph{Frequent retrain} 

In the original \vbmc algorithm, the GP model and variational posterior are re-trained only at the end of each iteration, corresponding to $n_\text{active} = 5$ likelihood evaluations. However, in the presence of observation noise, approximation of both the GP and the variational posterior may benefit from a more frequent update. 
Thus, for noisy likelihoods we introduced a \emph{frequent retrain}, that is fast re-training of both the GP and of the variational posterior within the active sampling loop, after each new function evaluation.
This frequent update sets \vbmc on par with other algorithms, such as \gpimiqr and \wsabi, which similarly retrain the GP representation after each likelihood evaluation.
In \vbmc, frequent retrain is active throughout the warm-up stage. After warm-up, we activate frequent retrain only when the previous iteration's reliability index $r(t-1) > 3$, indicating that the solution has not stabilized yet.

\subsection{Variational whitening}
\label{sec:supp_varwhit}

We start performing \emph{variational whitening} $\tau_\text{vw}$ iterations after the end of warm-up, and then subsequently at increasing intervals of $k \tau_\text{vw}$ iterations, where $k$ is the count of previously performed whitenings ($\tau_\text{vw} = 5$ in this work). Moreover, variational whitening is postponed until the reliability index $r(t)$ of the current iteration is below 3, indicating a degree of stability of the current variational posterior (see Section \ref{sec:features}).
Variational whitening consists of a linear transformation $\mat{W}$ of the inference space (a rotation and rescaling) such that the variational posterior $q_{\qparams}$ obtains unit diagonal covariance matrix. We compute the covariance matrix $\mat{C}_{\qparams}$ of $q_{\qparams}$ analytically, and we set the entries whose correlation is less than 0.05 in absolute value to zero, yielding a corrected covariance matrix $\widetilde{\mat{C}}_{\qparams}$. We then calculate the whitening transform $\mat{W}$ by performing a singular value decomposition (SVD) of $\widetilde{\mat{C}}_{\qparams}$.

\section{Acquisition functions}
\label{sec:supp_acqfuns}

In this Section, we report derivations and additional implementation details for the acquisition functions introduced in the main text.

\subsection{Observation noise}

All acquisition functions in the main text require knowledge of the log-likelihood observation noise $\sigman(\vtheta)$ at an arbitrary point $\vtheta \in \X$. However, we only assumed availability of an estimate $(\widehat{\sigma}_\text{obs})_n$ of $\sigman(\x_n)$ for all parameter values evaluated so far, $1 \le n \le N$.
We estimate values of $\sigman(\x)$ outside the training set via a simple nearest-neighbor approximation, that is
\begin{equation} \label{eq:kmeans}
\sigman(\x_\star) = \sigman(\x_n) \quad \text{ for } \; n = \arg\min_{1 \le n \le N} d_{\bm{\ell}}(\x_\star, \x_n),
\end{equation}
where $d_{\bm{\ell}}$ is the rescaled Euclidean distance between two points in inference space, where each coordinate dimension $i$ has been rescaled by the GP input length $\ell_i$, with $1 \le i \le \nparams$. When multiple GP hyperparameter samples are available, we use the geometric mean of each input length across samples. Eq. \ref{eq:kmeans} may seem like a coarse approximation, but we found it effective in practice.

\subsection{Expected information gain (\EIG)}

The \emph{expected information gain} (\EIG) acquisition function $a_\text{EIG}$ is based on a mutual information maximizing acquisition function for Bayesian quadrature introduced in \cite{gessner2019active}.

First, note that the \emph{information gain} is defined as the KL-divergence between posterior and prior; in our case, between the posterior  of the log joint $\logjoint$ after observing value $y_\star$ at $\x_\star$, and the current posterior over $\logjoint$ given the observed points in the training set, $\gpdata = \left\{\xx, \y, \bm{\sigma}_\text{obs} \right\}$. Since $y_\star$ is yet to be observed, we consider then the \emph{expected} information gain of performing a measurement at $\x_\star$, that is
\begin{equation} \label{eq:eigkl}
\text{EIG}(\x_\star; \gpdata_t) = \mathbb{E}_{y_\star | \x_\star} \left[D_{\text{KL}}\left(p(\logjoint | \gpdata \cup \left\{ \left(\x_\star, y_\star, {\sigman}_\star\right) \right\}) \; || \; p(\logjoint | \gpdata) \right)\right].
\end{equation}
It can be shown that Eq. \ref{eq:eigkl} is identical to the \emph{mutual information} between $\logjoint$ and $\y_\star$ \cite{mackay2003information}
\begin{equation}
I\left[\logjoint; y_\star\right] = H[\logjoint] + H[y_\star] - H[\logjoint,y_\star]
\end{equation}
where $H(\cdot)$ denotes the (marginal) differential entropy and $H(\cdot,\cdot)$ the joint entropy.
By the definition of GP, $y_\star$ is normally distributed, and so is each component $\logjoint_k$ of the log-joint, due to Bayesian quadrature (see Section \ref{sec:background}). As a weighted sum of normally distributed random variables, $\logjoint$ is also normally distributed, and so is the joint distribution of $y_\star$ and $\logjoint$.
We recall that the differential entropy of a bivariate normal distribution with covariance matrix $\mat{A} \in \mathbb{R}^{2 \times 2}$ is $H = \log(2\pi e) + \frac{1}{2} \log |\mat{A}|$. Thus we have (see Eq. \ref{eq:eig} in the main text)
\begin{equation}
a_\text{EIG}(\x_\star) \equiv I\left[\logjoint; y_\star\right] = -\frac{1}{2} \log \left(1 - \rho^2(\x_\star) \right), \qquad \text{with} \;
\rho(\x_\star) \equiv \frac{\mathbb{E}_{\qparams}\left[ C_\gpdata(f(\cdot), f(\x_\star)) \right]}{\sqrt{v_{\gpdata}(\x_\star) \mathbb{V}_{f|\gpdata}[\logjoint]}},
\end{equation}
where we used the \emph{scalar correlation} $\rho(\cdot)$ \cite{gessner2019active}; and $C_{\gpdata}(\cdot,\cdot)$ is the GP posterior covariance, $v_{\gpdata}(\cdot)$ the GP posterior predictive variance (including observation noise), and $\mathbb{V}_{f|\gpdata}[\logjoint]$ the posterior variance of the expected log joint -- all given the current training set $\gpdata$.

The expected value at the numerator of $\rho(\x_\star)$ is
\begin{equation} \label{eq:covint}
\begin{split}
\mathbb{E}_{\qparams}\left[ C_\gpdata(f(\cdot), f(\x_\star)) \right] = & \, 
\int q(\x) C_{\gpdata}\left(f(\x),f(\x_\star)\right)  \, d\x \\
 = & \, \sum_{k = 1}^K w_k \int \normpdf{\x}{\vmu_k}{\sigma_k^2 \bm{\Sigma}} C_{\gpdata}\left(f(\x),f(\x_\star)\right) \, d\x \\
 = & \, \sum_{k = 1}^K w_k \mathcal{K}_k(\x_\star),
\end{split}
\end{equation}
where we recall that $w_k$, $\mmu_k$, and $\sigma_k$ are, respectively, the mixture weight, mean, and scale of the $k$-th component of the variational posterior $q$, for $1 \le k \le K$; $\mathbf{\Sigma}$  is a common diagonal covariance matrix $\mathbf{\Sigma} \equiv \text{diag}[{{\lambda}^{(1)}}^2,\ldots,{\lambda^{(\nparams)}}^2]$; and $C_{\gpdata}$ is the GP posterior covariance as per Eq. \ref{eq:postgp}.
Finally, each term in Eq. \ref{eq:covint} can be written as
\begin{equation} \label{eq:covint1}
\begin{split}
\mathcal{K}_{k}(\x_\star) = & \int \normpdf{\x}{\vmu_k}{\sigma_k^2 \bm{\Sigma}} \Big[ \sigma_f^2 \Lambda \normpdf{\x}{\x_\star}{\bm{\Sigma}_\ell} \ldots \\ 
 & \ldots - \sigma_f^2 \Lambda \normpdf{\x}{\xx}{\bm{\Sigma}_\ell} \left[\kappa(\xx,\xx) + \bm{\Sigma}_\text{obs}(\xx)\right]^{-1} \sigma_f^2 \Lambda \normpdf{\xx}{\x_\star}{\bm{\Sigma}_\ell} \Big] d\x \\
  =  & \, \sigma_f^2 \Lambda \normpdf{\x_\star}{\vmu_k}{\bm{\Sigma}_\ell + \sigma_k^2 \bm{\Sigma}} - \sigma_f^2 \Lambda \z^\top_k \left[\kappa(\xx,\xx) + \bm{\Sigma}_\text{obs}(\xx)\right]^{-1} \normpdf{\xx}{\x_\star}{\bm{\Sigma}_\ell}, \\
 \end{split}
\end{equation}
where $\z_k$ is a $N$-dimensional vector with entries $z_k^{(n)} = \sigma_f^2 \Lambda \normpdf{\vmu_k}{\x_n}{\sigma_k^2 \bm{\Sigma} + \bm{\Sigma}_\ell}$ for $1 \le n \le N$.


\subsection{Integrated median / variational interquantile range (\IMIQR / \VIQR)}

The \emph{integrated median interquantile range} (\imiqr) acquisition function has been recently proposed in \cite{jarvenpaa2020parallel} as a robust, principled metric for posterior estimation with noisy evaluations (see also Eq. \ref{eq:imiqr} in the main text),
\begin{equation} \label{eq:imiqrapp}
a_\text{IMIQR}(\x_\star) = - 2 \int_{\X} \exp\left({\overline{f}_{\gpdata}(\x)}\right) \sinh\left(u s_{\gpdata \cup \x_\star}(\x) \right) d\x.
\end{equation}
It combines two ideas: (a) using the interquantile range (\iqr) as a robust measure of uncertainty, as opposed to the variance; and (b) approximating the median integrated \iqr loss, which follows from decision-theoretic principles but is intractable, with the integrated median \iqr, which can be computed somewhat more easily \cite{jarvenpaa2020parallel}. Note that Eq. \ref{eq:imiqrapp} differs slightly from Eq. 30 in \cite{jarvenpaa2020parallel} in that in our definition the prior term is subsumed into the joint distribution, with no loss of generality.

A major issue with Eq. \ref{eq:imiqrapp} is that the integral is still intractable. By noting that $\exp\left({\overline{f}_{\gpdata}(\x)}\right)$ is the joint distribution, in \vbmc we can replace it with the variational posterior, obtaining thus the \emph{variational} (integrated median) \emph{interquantile range} acquisition function $a_\text{VIQR}$ (see main text).

\section{Benchmark details}

We report here details about the benchmark setup, in particular parameter bounds and dataset information for all problems in the benchmark (Section \ref{sec:specs}); how we adapted the \wsabi and \gpimiqr algorithms for the purpose of our noisy benchmark (Section \ref{sec:algos}); and the computing infrastructure (Section \ref{sec:computing}).

\subsection{Problem specification}
\label{sec:specs}

\paragraph*{Parameter bounds}

We report in Table \ref{tab:params} the parameter bounds used in the problems of the noisy-inference benchmark. We denote with \lb and \ub the hard lower and upper bounds, respectively; whereas with \plb and \pub we denote the `plausible' lower and upper bounds, respectively \cite{acerbi2017practical,acerbi2018variational}. Plausible ranges should identify a region of high posterior probability mass in parameter space given our knowledge of the model and of the data; lacking other information, these are recommended to be set to e.g. the $\sim 68$\% high-density interval according to the marginal prior probability in each dimension \cite{acerbi2018variational}. Plausible values are used to initialize and set hyperparameters of some of the algorithms. For example, the initial design for \VBMC and \gpimiqr is drawn from a uniform distribution over the plausible box in inference space.

\begin{table}[htb]
\caption{Parameters and bounds of all models (before remapping to inference space).}
  \label{tab:params}
  \centering
\begin{tabular}{lclcccc}
    \toprule
Model & Parameter & Description & $\lb$ & $\ub$ & $\plb$ & $\pub$ \\
    \midrule
Ricker & $\log(r)$ & Growth factor (log) & 3 & 5 & 3.2 & 4.8 \\
& $\phi$ & Observed fraction & 4 & 20 & 5.6 & 18.4 \\
& $\sigma_\varepsilon$ & Growth noise & 0 & 0.8 & 0.08 & 0.72 \\
    \midrule
aDDM & $d$ & Drift rate & 0 & 5 & 0.1 & 2 \\
& $\beta$ & Attentional bias factor & 0 & 1 & 0.1 & 0.9 \\
& $\sigma_\varepsilon$ & Diffusion noise & 0.1 & 2 & 0.2 & 1 \\
& $\lambda$ & Lapse rate & 0.01 & 0.2 & 0.03 & 0.1 \\
    \midrule
Bayesian & $w_\text{s}$ & Sensory noise (Weber's fraction) & 0.01 & 0.5 & 0.05 & 0.25 \\
timing & $w_\text{m}$ & Motor noise (Weber's fraction) & 0.01 & 0.5 & 0.05 & 0.25 \\
& $\mu_\text{p}$ & Prior mean (seconds) & 0.3 & 1.95 & 0.6 & 0.975 \\
& $\sigma_\text{p}$ & Prior standard deviation (seconds) & 0.0375 & 0.75 & 0.075 & 0.375 \\
& $\lambda$ & Lapse rate & 0.01 & 0.2 & 0.02 & 0.05 \\
    \midrule
Multisensory & $\sigma_\text{vest}$ & Vestibular noise (deg) & 0.5 & 80 & 1 & 40 \\
causal & $\sigma_\text{vis}(c_\text{low})$ & Visual noise, low coherence (deg) & 0.5 & 80 & 1 & 40 \\
inference (CI) & $\sigma_\text{vis}(c_\text{med})$ & Visual noise, medium coherence (deg) & 0.5 & 80 & 1 & 40 \\
& $\sigma_\text{vis}(c_\text{high})$ & Visual noise, high coherence (deg) & 0.5 & 80 & 1 & 40 \\
& $\kappa$ & `Sameness' threshold (deg) & 0.25 & 180 & 1 & 45 \\
& $\lambda$ & Lapse rate & 0.005 & 0.5 & 0.01 & 0.2 \\
    \midrule
Neuronal & $\theta_1$ & Preferred direction of motion (deg) & 0 & 360 & 90 & 270 \\
selectivity & $\theta_2$ & Preferred spatial freq. (cycles/deg) & 0.05 & 15 & 0.5 & 10 \\
& $\theta_3$ & Aspect ratio of 2-D Gaussian & 0.1 & 3.5 & 0.3 & 3.2 \\
& $\theta_4$ & Derivative order in space & 0.1 & 3.5 & 0.3 & 3.2 \\
& $\theta_5$ & Gain inhibitory channel & -1 & 1 & -0.3 & 0.3 \\
& $\theta_6$ & Response exponent & 1 & 6.5 & 1.01 & 5 \\
& $\theta_7$ & Variance of response gain & 0.001 & 10 & 0.015 & 1 \\
    \midrule
Rodent & $w_0$ & Bias & -3 & 3 & -1 & 1 \\
2AFC & $w_\text{c}$ & Weight on `previous correct side' & -3 & 3 & -1 & 1 \\
 & $w_{\overline{s}}$ & Weight on long-term history & -3 & 3 & -1 & 1 \\
 & $w_\text{L}^{(0)}$ & Weight on left stimulus ($t=0$) & -3 & 3 & -1 & 1 \\
 & $w_\text{L}^{(-1)}$ & Weight on left stimulus ($t=-1$) & -3 & 3 & -1 & 1 \\
 & $w_\text{L}^{(-2)}$ & Weight on left stimulus ($t=-2$) & -3 & 3 & -1 & 1 \\
 & $w_\text{R}^{(0)}$ & Weight on right stimulus ($t=0$) & -3 & 3 & -1 & 1 \\
 & $w_\text{R}^{(-1)}$ & Weight on right stimulus ($t=-1$) & -3 & 3 & -1 & 1 \\
 & $w_\text{R}^{(-2)}$ & Weight on right stimulus ($t=-2$) & -3 & 3 & -1 & 1 \\
\bottomrule
\end{tabular}
\end{table}

\paragraph*{Dataset information}

\begin{itemize}
\item \textbf{Ricker:} We generated a synthetic dataset of $T = 50$ observations using the ``true'' parameter vector $\vtheta_\text{true} = (3.8,10,0.3)$ with $T = 50$, as in \cite{jarvenpaa2020parallel}.
\item \textbf{Attentional drift-diffusion model (aDDM):} We used fixation and choice data from two participants (subject \#13 and subject \#16 from \cite{krajbich2010visual}) who completed all $N = 100$ trials in the experiment without technical issues (reported as `missing trials' in the data).
\item \textbf{Bayesian timing:} We analyzed reproduced time intervals of one representative subject from Experiment 3 (uniform distribution; subject \#2) in \cite{acerbi2012internal}, with $N = 1512$ trials.
\item \textbf{Multisensory causal inference:} We examined datasets of subject \#1 and \#2 from the explicit causal inference task (`unity judgment') in \cite{acerbi2018bayesian}; with respectively $N = 1069$ and $N = 857$ trials, across three different visual coherence conditions.
\item \textbf{Neuronal selectivity:} We analyzed two neurons (one from area V1, one from area V2 of primate visual cortex) from \cite{goris2015origin}, both with $N = 1760$ trials. The same datasets have been used in previous optimization and inference benchmarks \cite{acerbi2017practical,acerbi2018variational,acerbi2019exploration}.
\item \textbf{Rodent 2AFC:} We took a representative rat subject from \cite{akrami2018posterior}, already used for demonstration purposes by \cite{roy2018efficient}, limiting our analysis of choice behavior to the last $N = 10^4$ trials in the data set.
\end{itemize}

\subsection{Algorithm specification}
\label{sec:algos}

\paragraph{\wsabi}

Warped sequential active Bayesian integration (\wsabi) is a technique to compute the log marginal likelihood via GP surrogate models and Bayesian quadrature \cite{gunter2014sampling}. In this work, we use \wsabi as an example of a surrogate-based method for model evidence approximation different from \VBMC. The key idea of \wsabi is to model directly the \emph{square root} of the likelihood function $\mathcal{L}$ (as opposed to the log-likelihood) via a GP,
\begin{equation} \label{eq:wsabi}
\tilde{\mathcal{L}}(\vtheta) \equiv \sqrt{2\left(\mathcal{L}(\vtheta) - \alpha \right)} \quad \Longrightarrow \quad \mathcal{L}(\vtheta) = \alpha + \frac{1}{2} \tilde{\mathcal{L}}(\vtheta)^2,
\end{equation}
where $\alpha$ is a small positive scalar. If $\tilde{\mathcal{L}}$ is modeled as a GP, $\mathcal{L}$ is not itself a GP (right-hand side of Eq. \ref{eq:wsabi}), but it can be approximated as a GP via a linearization procedure (\wsabil in \cite{gunter2014sampling}), which is the approach we follow throughout our work.

The \wsabi algorithm requires an unlimited inference space $\X \equiv \mathbb{R}^\nparams$ and a multivariate normal prior \cite{gunter2014sampling}. In our benchmark, all parameters have bound constraints, so we first map the original space to an unbounded inference space via a rescaled logit transform, with an appropriate log-Jacobian correction to the log posterior (see e.g., \cite{carpenter2017stan, acerbi2018variational}). Also, in our benchmark all priors are assumed to be uniform. Thus, we pass to \wsabi a `pseudo-prior' consisting of a multivariate normal centered on the middle of the \emph{plausible box}, and with standard deviations equal to half the \emph{plausible range} in each coordinate direction in inference space (see Section \ref{sec:specs}). We then correct for this added pseudo-prior by subtracting the log-pseudo-prior value from each log-joint evaluation.

\paragraph{\wsabi with noisy likelihoods}

The original \wsabi algorithm does not explicitly support observation noise in the (log-)likelihood. Thus, we modified \wsabi to include noisy likelihood evaluations, by mapping noise on the log-likelihood to noise in the square-root likelihood via an unscented transform, and by modifying \wsabi's uncertainty-sampling acquisition function to account for observation noise (similarly to what we did for $a_\text{npro}$, see main text). However, we found the noise-adjusted \wsabi to perform abysmally, even worse than the original \wsabi on our noisy benchmark. This failure is likely due to the particular representation used by \wsabi (Eq. \ref{eq:wsabi}). Crucially, even moderate noise on the log-likelihood translates to extremely large noise on the (square-root) likelihood. Due to this large observation noise, the latent GP will revert to the GP mean function, which corresponds to the constant $\alpha$ (Eq. \ref{eq:wsabi}). In the presence of modeled log-likelihood noise, thus, the GP representation of \wsabi becomes near-constant and practically useless. For this reason, here and in the main text we report the results of \wsabi \emph{without} explicitly added support for observation noise. More work is needed to find an alternative representation of \wsabi which would not suffer from observation noise, but it is beyond the scope of our paper.

\paragraph{\gpimiqr}

For the \gpimiqr algorithm described in \cite{jarvenpaa2020parallel}, we used the latest implementation (\textsc{v3}) publicly available at: \url{https://github.com/mjarvenpaa/parallel-GP-SL}. We considered the \imiqr acquisition function with sequential sampling strategy; the best-performing acquisition function in the empirical analyses in \cite{jarvenpaa2020parallel}. We used the code essentially `as is', with minimal changes to interface the algorithm to our noisy benchmark. We ran the algorithm with the recommended default hyperparameters. Given the particularly poor performance of \gpimiqr on some problems (e.g., Timing, Neuronal), which we potentially attributed to convergence failures of the default \MCMC sampling algorithm (\textsc{dram}; \cite{haario2006dram}), we also reran the method with an alternative and robust sampling method (parallel slice sampling; \cite{neal2003slice,acerbi2018bayesian}). However, performance of \gpimiqr with slice sampling was virtually identical, and similarly poor, to its performance with \textsc{dram} (data not shown). We note that the same grave issues with the Neuronal model emerged even when we forced initialization of the algorithm in close vicinity of the mode of the posterior (data not shown).
We attribute the inability of \gpimiqr to make significant progress on some problems to excessive exploration, which may lead to GP instabilities; although further investigation is needed to identify the exact causes, beyond the scope of this work.

\subsection{Computing infrastructure}
\label{sec:computing}

All benchmark runs were performed on \textsc{Matlab} 2017a (Mathworks, Inc.) using a High Performance Computing cluster whose details can be found at the following link: \url{https://wikis.nyu.edu/display/NYUHPC/Clusters+-+Prince}.
Since different runs may have been assigned to compute nodes with vastly different loads or hardware, we regularly assessed execution speed by performing a set of basic speed benchmark operations (\texttt{bench} in \textsc{Matlab}; considering only numerical tasks). Running times were then converted to the estimated running time on a reference machine, a laptop computer with 16.0 GB RAM and Intel(\textsc{r}) Core(\textsc{tm}) i7-6700\textsc{hq} CPU @ 2.60 GHz, forced to run single-core during the speed test.

\section{Additional results}

We include here a series of additional experimental results and plots omitted from the main text for reasons of space. First, we report the results of the posterior inference benchmark with a different metric (Section \ref{sec:supp_gskl}). Then, we present results of a robustness analysis of solutions across runs (Section \ref{sec:supp_worse}) and of an ablation study (Section \ref{sec:supp_ablation}). In Section \ref{sec:posteriors}, we show a comparison of true and approximate posteriors for all problems in the benchmark. 
Then, we study sensitivity of \vbmc-\viqr to imprecision in the log-likelihood noise estimates (Section \ref{sec:noisenoise}).
 Finally, we report results for an additional synthetic problem, the g-and-k model (Section \ref{sec:gandk}).

\subsection{Gaussianized symmetrized KL divergence (gsKL) metric}
\label{sec:supp_gskl}

In the main text, we measured the quality of the posterior approximation via the mean marginal total variation distance (MMTV) between true and appoximate posteriors, which quantifies the distance between posterior marginals. Here we consider an alternative loss metric, the ``Gaussianized'' symmetrized Kullback-Leibler divergence (gsKL), which is sensitive instead to differences in means and covariances \cite{acerbi2018variational}. Specifically, the gsKL between two pdfs $p$ and $q$ is defined as
\begin{equation} \label{eq:gsKL}
\text{gsKL}(p, q) = \frac{1}{2}\left[D_\text{KL}\left(\mathcal{N}[p]||\mathcal{N}[q]\right) + D_\text{KL}(\mathcal{N}[q]|| \mathcal{N}[p])\right],
\end{equation}
where $\mathcal{N}[p]$ is a multivariate normal distribution with mean equal to the mean of $p$ and covariance matrix equal to the covariance of $p$ (and same for $q$). Eq. \ref{eq:gsKL} can be expressed in closed form in terms of the means and covariance matrices of $p$ and $q$.

\begin{figure}[htb]
  \includegraphics[width=\linewidth]{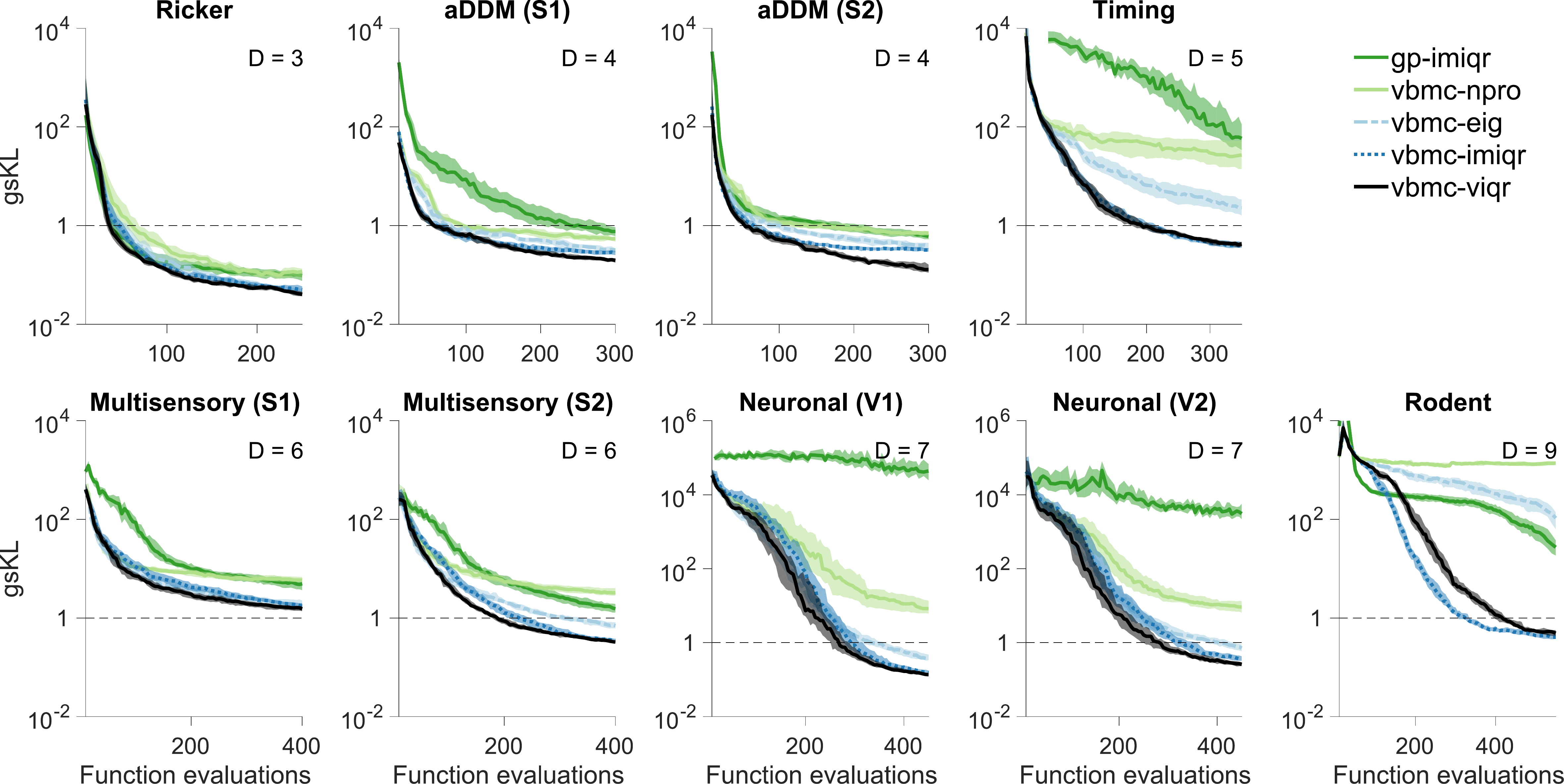}  
\vspace{-1em}
\caption{{\bf Posterior estimation loss (gsKL).} Median Gaussianized symmetrized KL divergence (gsKL) between the algorithm's posterior and ground truth, as a function of number of likelihood evaluations. A desirable target (dashed line) is less than 1. Shaded areas are 95\% CI of the median across 100 runs.
}
  \label{fig:gsKL}
\end{figure}

Fig. \ref{fig:gsKL} shows the gsKL between approximate posterior and ground truth, for all algorithms and inference problems considered in the main text. 
For reference, two Gaussians with unit variance and whose means differ by $\sqrt{2}$ (resp., $\frac{1}{2}$) have a gsKL of 1 (resp., $\frac{1}{8}$). For this reason, we consider a desirable target to be (much) less than 1. Results are qualitatively similar to what we observed for the MMTV metric (Fig. \ref{fig:mmtv} in the main text), in that the ranking and convergence properties of different methods is the same for MMTV and gsKL. In particular, previous state-of-the art method \gpimiqr fails to converge in several challenging problems (Timing, Neuronal and Rodent); among the variants of \VBMC,  \vbmc-\viqr and \vbmc-\imiqr are the only ones that perform consistently well.

\subsection{Worse-case analysis (90\% quantile)}
\label{sec:supp_worse}

In the main text and other parts of this \supplement, we showed for each performance metric the \emph{median} performance across multiple runs, to convey the `average-case' performance of an algorithm; in that we expect performance to be at least as good as the median for about half of the runs. To assess the robustness of an algorithm, we are also interested in a `worse-case' analysis that looks at higher quantiles of the distribution of performance, which are informative of how bad performance can reasonably get (e.g., we expect only about one run out of ten to be worse than the 90\% quantile).

\begin{figure}[htb]
  \includegraphics[width=\linewidth]{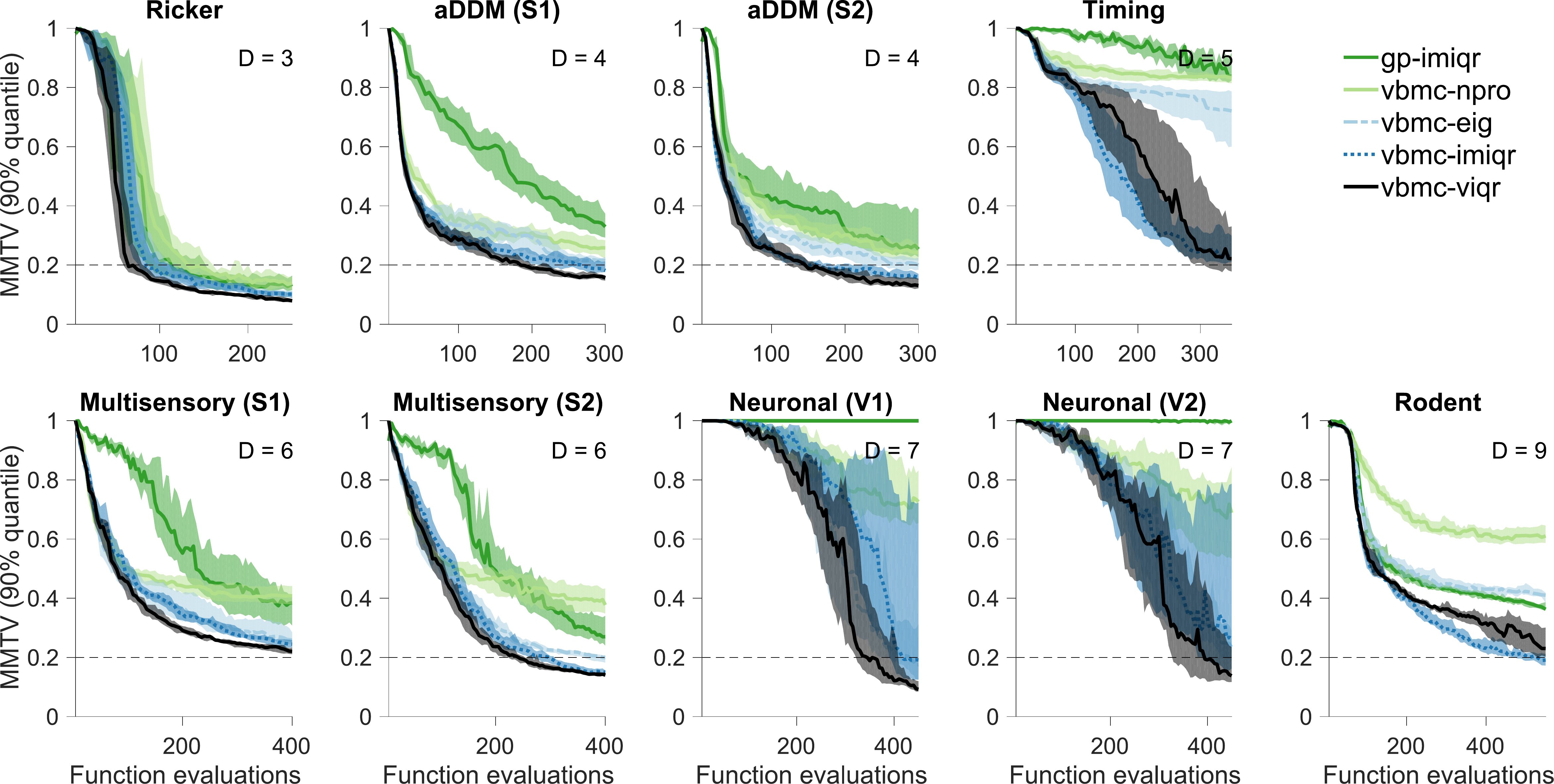}  
\vspace{-1.25em}
\caption{{\bf Worse-case posterior estimation loss (MMTV).} 90\% quantile of the mean marginal total variation distance (MMTV) between the algorithm's posterior and ground truth, as a function of number of likelihood evaluations. A desirable target (dashed line) is less than 0.2, corresponding to more than 80\% overlap between true and approximate posterior marginals (on average across model parameters). Shaded areas are 95\% CI of the 90\% quantile across 100 runs.
}
  \label{fig:mmtv90}
\end{figure}

We show in Figure \ref{fig:mmtv90} the 90\% quantile of the MMTV metric, to be compared with Fig. \ref{fig:mmtv} in the main text (results for other metrics are analogous). These results show that the best-performing algorithms, 
\vbmc-\viqr and \vbmc-\imiqr, are also the most robust, as both methods manage to achieve good solutions most of the time (with one method working slightly better than the other on some problems, and vice versa). 
By contrast, other methods such as \gpimiqr show more variability, in that on some problems (e.g., aDDM) they may have reasonably good median performance, but much higher error when looking at the 90\% quantile.

\subsection{Ablation study}
\label{sec:supp_ablation}

We show here the performance of the \vbmc algorithm after removing some of the features considered in the main paper. As a baseline algorithm we take \vbmc-\viqr. First, we show \vbmc-\textsc{nowv}, obtained by removing from the baseline the `variational whitening' feature (see main text and Section \ref{sec:supp_varwhit}). Second, we consider a variant of \vbmc-\viqr in which we do not sample GP hyperparameters from the hyperparameter posterior, but simply obtain a point estimate through maximum-a-posteriori estimation (\vbmc-\textsc{map}). Optimizing GP hyperperameters, as opposed to a Bayesian treatment of hyperparameters, is a common choice for many surrogate-based methods (e.g., \wsabi, \gpimiqr, although the latter integrates analytically over the GP mean function), so we investigate whether it is needed for \vbmc.
Finally, we plot the performance of \vbmc in its original implementation (\vbmc-\textsc{old}), as per the \vbmc paper \cite{acerbi2018variational}. For reference, we also plot both the \vbmc-\viqr and \gpimiqr algorithms, as per Fig. \ref{fig:mmtv} in the main text. 

\begin{figure}[htb]
  \includegraphics[width=\linewidth]{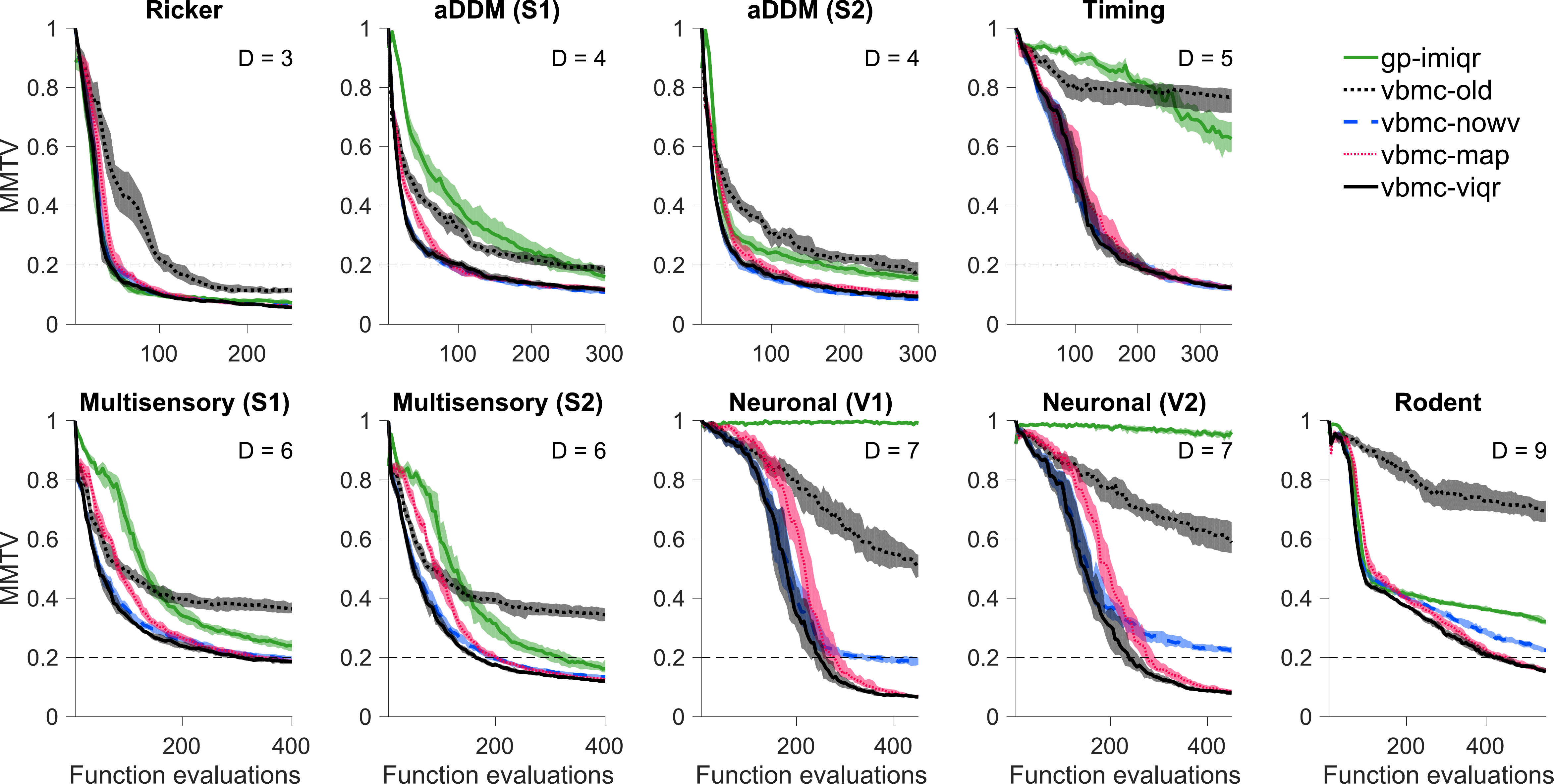}  
\vspace{-1.25em}
\caption{{\bf Lesion study; posterior estimation loss (MMTV).} Median mean marginal total variation distance (MMTV) between the algorithm's posterior and ground truth, as a function of number of likelihood evaluations. Shaded areas are 95\% CI of the median across 100 runs.
}
  \label{fig:variants}
\end{figure}

We show in Fig. \ref{fig:variants} the results for the MMTV metric, although results are similar for other inference metrics. We can see that all `lesioned' versions of \vbmc perform generally worse than \vbmc-\viqr, to different degree, and more visibly in more difficult inference problems. However, for example, \vbmc-\textsc{map} still performs substantially better than \gpimiqr, suggesting that the difference in performance between \vbmc and  \gpimiqr is not simply because \vbmc marginalizes over GP hyperparameters. It is also evident that the previous version of \vbmc (\vbmc-\textsc{old}) is extremely ineffective in the presence of noisy log-likelihoods.

\subsection{Comparison of true and approximate posteriors}
\label{sec:posteriors}

\begin{figure}[htb!]
\includegraphics[width=\linewidth]{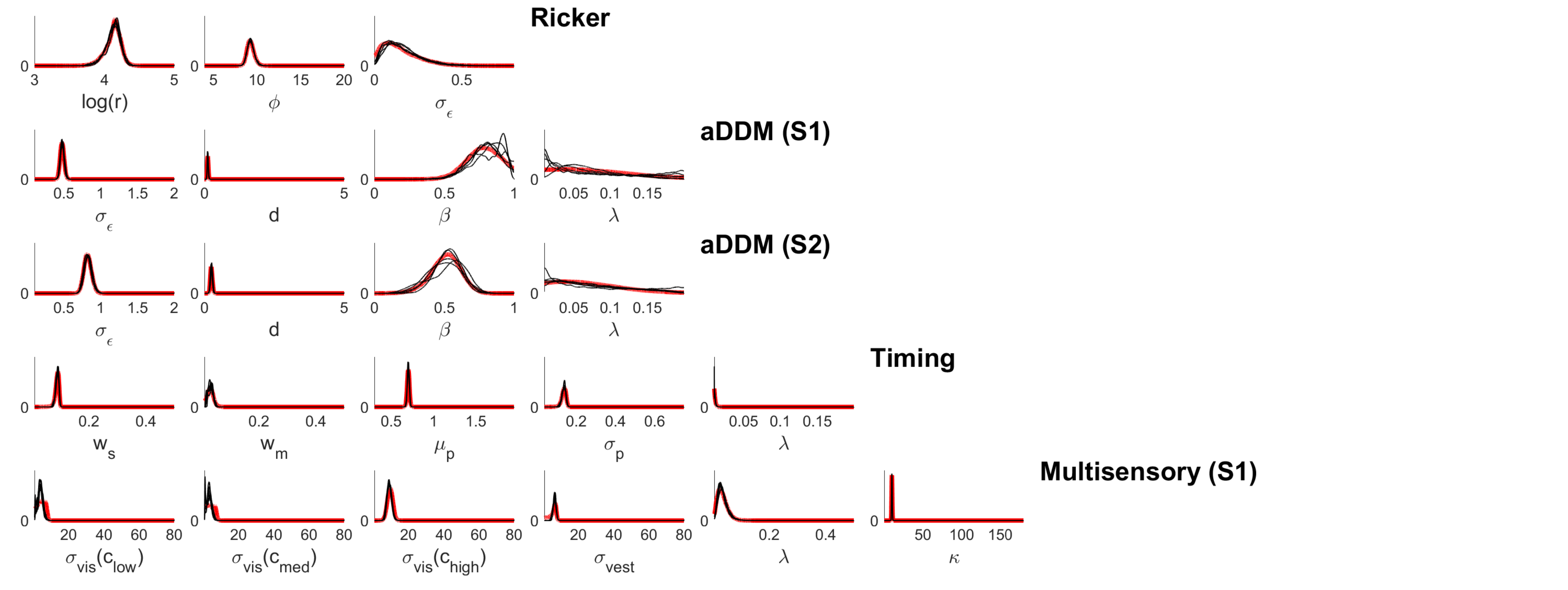}  
\includegraphics[width=\linewidth]{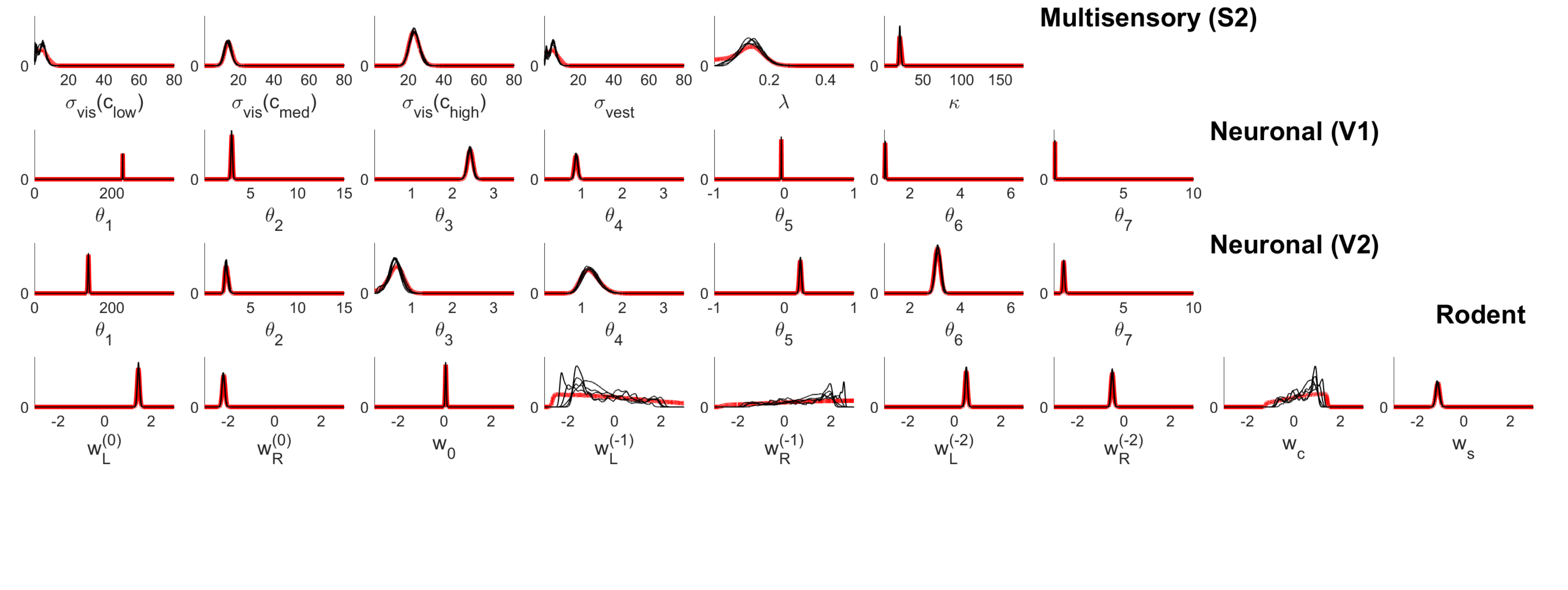}  
\vspace{-3.75em}
  \caption{{\bf True and approximate marginal posteriors.} Each panel shows the ground-truth marginal posterior distribution (red line) for each parameter of problems in the noisy benchmark (rows). For each problem, black lines are marginal distributions of five randomly-chosen approximate posteriors returned by \VBMC-\viqr.
}
  \label{fig:post}
\end{figure}

We plot in Fig. \ref{fig:post} a comparison between the `true' marginal posteriors, obtained for all problems via extensive \MCMC sampling, and example approximate posteriors recovered by \vbmc-\viqr after $50\times(D+2)$ likelihood evaluations, the budget allocated for the benchmark.
As already quantified by the MMTV metric, we note that \VBMC is generally able to obtain good approximations of the true posterior marginals. The effect of noise becomes more prominent when the posteriors are nearly flat, in which case we see greater variability in the \vbmc solutions for some parameters (e.g., in the challenging Rodent problem). Note that this is also a consequence of choosing non-informative uniform priors over bounded parameter ranges in our benchmark, which is not necessarily best practice on real problems; (weakly) informative priors should be preferred in most cases \cite{gelman2013bayesian}.

\begin{figure}[htb!]
  \includegraphics[width=0.5\linewidth]{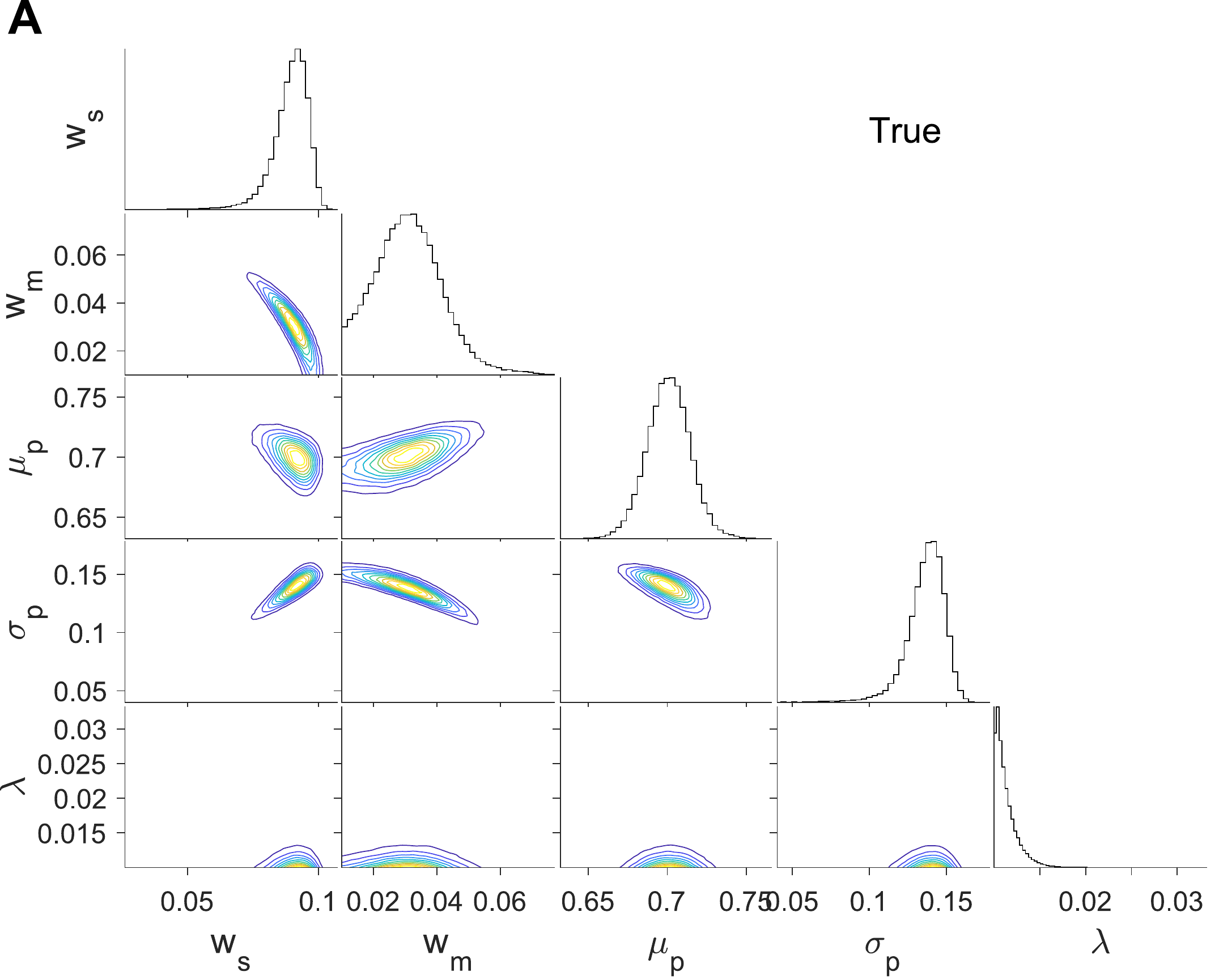}  
  \includegraphics[width=0.5\linewidth]{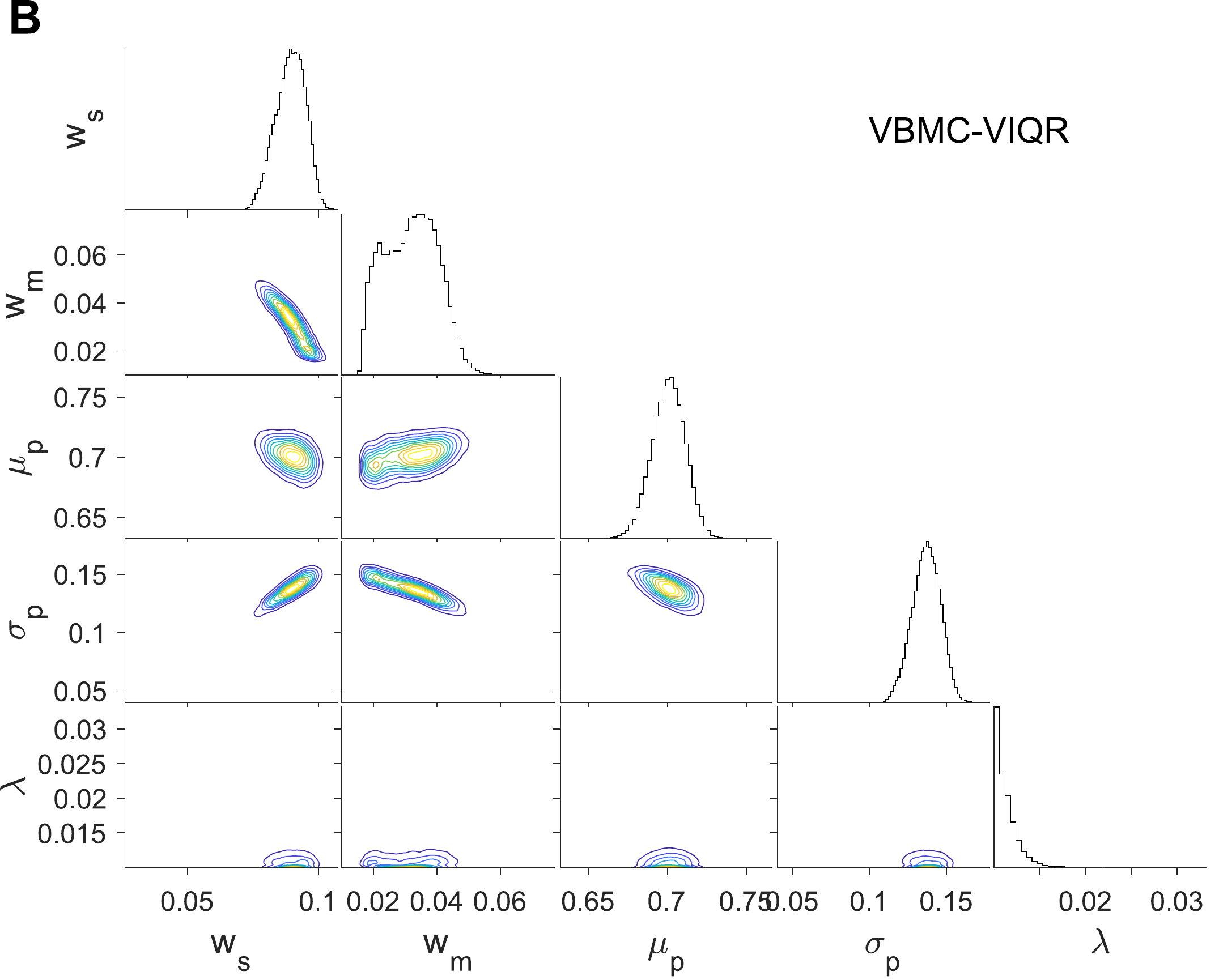}  
\vspace{-1em}
  \caption{{\bf True and approximate posterior of Timing model.} \textbf{A.} Triangle plot of the `true' posterior (obtained via \MCMC) for the Timing model. Each panel below the diagonal is the contour plot of the 2-D marginal posterior distribution for a given parameter pair. Panels on the diagonal are histograms of the 1-D marginal posterior distribution for each parameter (as per Fig. \ref{fig:post}). \textbf{B.} Triangle plot of a typical variational solution returned by \VBMC-\viqr.
}
  \label{fig:timingpost}
\end{figure}

To illustrate the ability of \vbmc-\viqr to recover complex interactions in the posterior distribution (and not only univariate marginals) in the presence of noise, we plot in Fig. \ref{fig:timingpost} the full pairwise posterior for one of the problems in the benchmark (Timing model). We can see that the approximate posterior matches the true posterior quite well, with some underestimation of the distribution tails. Underestimation of posterior variance is a common problem for variational approximations \cite{blei2017variational} and magnified here by the presence of noisy log-likelihood evaluations, and it represents a potential direction of improvement for future work.

\subsection{Sensitivity to imprecise noise estimates}
\label{sec:noisenoise}

In this paragraph, we look at how robust \vbmc-\viqr is to different degrees of imprecision in the estimation of log-likelihood noise. We consider the same setup with three example problems as in the noise sensitivity analysis reported in main text (Fig. \ref{fig:noise} in the main text). For this analysis, we fixed the emulated noise to $\sigman(\x) = 2$ for all problems. We then assumed that the estimated noise $\widehat{\sigma}_\text{obs}(\x)$, instead of being known (nearly) exactly, is drawn randomly as $\widehat{\sigma}_\text{obs} \sim \text{Lognormal}\left(\ln \sigma_\text{obs}, \sigma_\sigma^2 \right)$, where $\sigma_\sigma \ge 0$ represents the jitter of the noise estimates on a logarithmic scale.

\begin{figure}[htb]
\includegraphics[width=\linewidth]{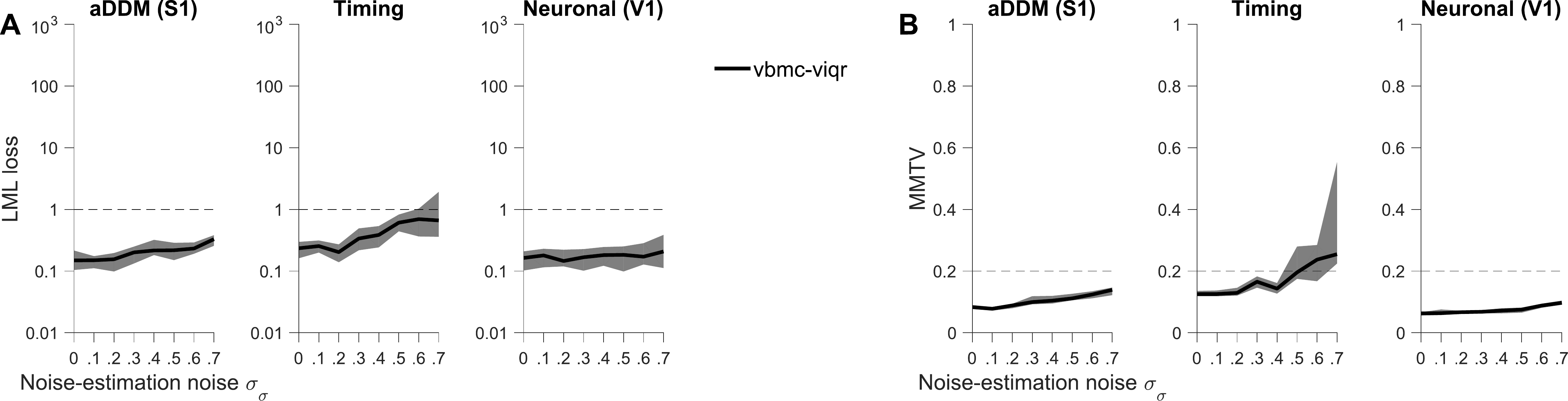}
\vspace{-1em}
\caption{{\bf Sensitivity to imprecise noise estimates.} Performance metrics of \vbmc-\viqr with respect to ground truth, as a function of noise-of-estimating-noise $\sigma_\sigma$. For all metrics, we plot the median and shaded areas are 95 \% CI of the median across 50 runs. \textbf{A.} Absolute error of the log marginal likelihood (LML) estimate. \textbf{B.} Mean marginal total variation distance (MMTV).}
  \label{fig:noisenoise}
\end{figure}

We tested the performance of \vbmc-\viqr for different values of noise-of-estimating-noise, $\sigma_\sigma \ge 0$ (see Fig. \ref{fig:noisenoise}). We found that up to $\sigma_\sigma \approx 0.4$ (that is, $\widehat{\sigma}_\text{obs}$ varying roughly between $0.5-2.2$ times the true value) the quality of the inference degrades only slightly. For example, at worst the \textsc{mmtv} metric rises from $0.13$ to $0.16$ on the Timing problem (less than $\sim 25\%$ increase), and in the other problems it is barely affected. These results show that \vbmc-\viqr is quite robust to imprecise noise estimates. Combined with the fact that we expect estimates of the noise obtained from methods such as \textsc{ibs} to be very precise \cite{van2020unbiased}, imprecision in the noise estimates should not be an issue in practice.

\subsection{g-and-k model}
\label{sec:gandk}

We report here results of another synthetic test model omitted from the main text.
The g-and-k model is a common benchmark simulation model represented by a flexible probability distribution defined via its quantile function,
\begin{equation} \label{eq:gandk}
Q\left(\Phi^{-1}(p); \vtheta \right) = a + b \left(1 + c \frac{1 - \exp\left(-g \Phi^{-1}(p)\right)}{1 + \exp\left(-g \Phi^{-1}(p)\right)} \right) \left[1 + \left(\Phi^{-1}(p) \right)^2 \right]^k \Phi^{-1}(p),
\end{equation}
where $a,b,c,g$ and $k$ are parameters and $p \in [0,1]$ is a quantile.
As in previous studies, we fix $c = 0.8$ and infer the parameters $\vtheta = (a,b,g,k)$ using the synthetic likelihood (SL) approach \cite{wood2010statistical,price2018bayesian,jarvenpaa2020parallel}. 
We use the same dataset as \cite{price2018bayesian,jarvenpaa2020parallel}, generated with ``true'' parameter vector $\vtheta_\text{true} = (3,1,2,0.5)$, and for the log-SL estimation the same four summary statistics obtained by fitting a skew $t$-distribution to a set of samples generated from Eq. \ref{eq:gandk}.
We use $N_\text{sim} = 100$, which produces fairly precise observations, with $\sigman(\vtheta_\text{MAP}) \approx 0.14$. In terms of parameter bounds, we set $\lb = (2.5,0.5,1.5,0.3)$ and $\ub = (3.5,1.5,2.5,0.7)$ as in \cite{jarvenpaa2020parallel}; and $\plb = (2.6,0.6,1.6,0.34)$ and $\pub = (3.4,1.4,2.4,0.66)$.

\begin{figure}[htb]
  \includegraphics[width=\linewidth]{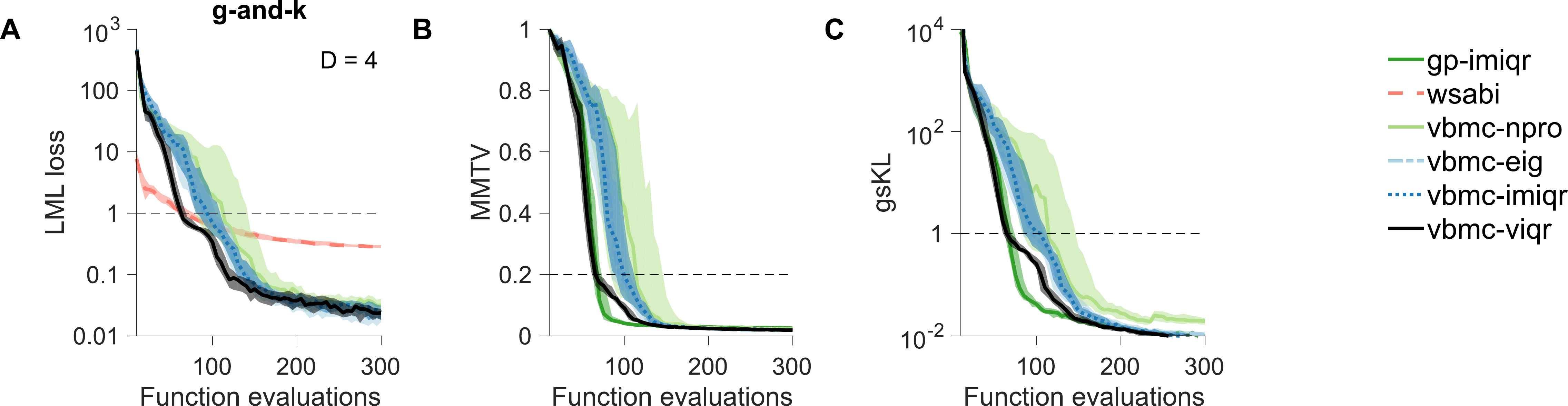}  
\vspace{-0.5em}
\caption{{\bf Performance on g-and-k model.} Performance metrics of various algorithms with respect to ground truth, as a function of number of likelihood evaluations, on the g-and-k model problem. For all metrics, we plot the median and shaded areas are 95\% CI of the median across 100 runs. \textbf{A.} Absolute error of the log marginal likelihood (LML) estimate. \textbf{B.} Mean marginal total variation distance (MMTV). \textbf{C.} ``Gaussianized'' symmetrized Kullback-Leibler divergence (gsKL).}
  \label{fig:gandk}
\end{figure}

We show in Fig. \ref{fig:gandk} the performance of all methods introduced in the main text for three different inference metric: the log marginal likelihood (LML) loss, and both the mean marginal total variation distance (MMTV) and the ``Gaussianized'' symmetrized Kullback-Leibler divergence (gsKL) between approximate posterior and ground-truth posterior. For algorithms other than VBMC, we only report metrics they were designed for (posterior estimation for \gpimiqr, model evidence for \wsabi). The plots show that almost all algorithms (except \wsabi) eventually converge to a very good performance across metrics, with only some differences in the speed of convergence. These results suggest that the g-and-k problem as used, e.g., in \cite{jarvenpaa2020parallel} might be a relatively easy test case for surrogate-based Bayesian inference; as opposed to the challenging real scenarios of our main benchmark, in which we find striking differences in performance between algorithms. Since we already present a simple synthetic scenario in the main text (the Ricker model), we did not include the g-and-k model as part of our main noisy-benchmark.

Finally, we note that when performing simulation-based inference based on summary statistics (such as here with the g-and-k model, and the Ricker model discussed in the main text), computing the marginal likelihood may not be a reliable approach for model comparison \cite{robert2011lack}. However, this is not a concern when performing simulation-based inference with methods that compute the log-likelihood with the entire data, such as \textsc{ibs} \cite{van2020unbiased}, as per all the other example problems in the main text.

\end{document}